%% file: main.tex
\documentclass{article}

\usepackage[preprint]{neurips_2024}

\usepackage[utf8]{inputenc} 
\usepackage[T1]{fontenc}    
\usepackage{hyperref}       
\usepackage{url}            
\usepackage{booktabs}       
\usepackage{amsfonts}       
\usepackage{nicefrac}       
\usepackage{microtype}      
\usepackage{xcolor}         

\usepackage{bm}
\usepackage{wrapfig}
\usepackage[bottom]{footmisc}
\usepackage{amsmath}
\usepackage{amssymb}
\usepackage{mathtools}
\usepackage{amsthm}
\usepackage[capitalize,noabbrev]{cleveref}

\DeclareMathOperator*{\argmax}{arg\,max}

\theoremstyle{plain}
\newtheorem{theorem}{Theorem}[section]
\newtheorem{proposition}[theorem]{Proposition}
\newtheorem*{proposition*}{Proposition}

\newtheorem*{fact*}{Fact}

\theoremstyle{definition}

\theoremstyle{remark}

\title{On Stateful Value Factorization\\in Multi-Agent Reinforcement Learning}

\newcommand*\samethanks[1][\value{footnote}]{\footnotemark[#1]}

\author{%
    Enrico Marchesini\thanks{Contact authors: emarche@mit.edu, c.amato@northeastern.edu} \\
    LIDS \\
    Massachusetts Institute of Technology \\
    Cambridge (MA), USA \\
\And
    Andrea Baisero, Rupali Bhati, Christopher Amato\samethanks \\
    Khoury College of Computer Sciences \\
    Northeastern University \\
    Boston (MA), USA
}

\begin{document}

\maketitle

\input{Sections/abstract}
\input{Sections/introduction}

\input{Sections/preliminaries}
\input{Sections/methods}
\input{Sections/experiments}
\input{Sections/conclusions}

\bibliographystyle{apalike}
\bibliography{biblio}



\appendix
\input{Sections/appendices}

\end{document}

%% file: Sections/abstract.tex
\begin{abstract}
Value factorization is a popular paradigm for designing scalable multi-agent reinforcement learning algorithms. However, current factorization methods make choices without full justification that may limit their performance. For example, the theory in prior work uses stateless (i.e., history) functions, while the practical implementations use state information---making the motivating theory a mismatch for the implementation. Also, methods have built off of previous approaches, inheriting their architectures without exploring other, potentially better ones. To address these concerns, we formally analyze the theory of using the state instead of the history in current methods---reconnecting theory and practice. We then introduce DuelMIX, a factorization algorithm that learns distinct per-agent utility estimators to improve performance and achieve full expressiveness. Experiments on StarCraft II micromanagement and Box Pushing tasks demonstrate the benefits of our intuitions. 
\end{abstract}

%% file: Sections/introduction.tex
\section{Introduction}
\label{sec:introduction}

Recent advancements in multi-agent reinforcement learning (MARL) have led to impressive results in many complex cooperative tasks \citep{starcraft, mpe, alp2023mrs}. 
Many of these MARL methods use
\textit{Centralised Training with Decentralised Execution} (CTDE) \citep{maddpg}, which allows them to train in a centralized fashion but still execute in a decentralized manner.

The dominant form of CTDE in value-based MARL is value factorization \citep{vdn, qmix, wqmix, qplex, gdq, marche22marl}.
These methods factor a (centralized) joint action value into (decentralized) per-agent utilities conditioned on local information. The resulting approaches ensure the greedy action selection from each agent's local utility is the same as greedy action selection over the centralized value function (i.e., the argmax over the local utilities is the same as a joint argmax over the centralized value function)---the \textit{Individual Global Max} (IGM) principle \citep{qtran}. IGM provides decentralization and scalability since algorithms no longer need to perform costly joint maximization over all agents.  
Earlier forms of factorization used strong constraints to ensure IGM (e.g., linearly or monotonically combining the local utilities in VDN \citep{vdn} and QMIX \citep{qmix}) which limited their expressiveness. More recent methods such as QPLEX \citep{qplex} can, in theory, represent the full set of IGM factorizations.

Despite the merits of value factorization, there is a mismatch between the theory and practice of these methods. In particular, the theory behind the methods assumes history information at the local and centralized levels while most practical implementations replace history with (ground truth) state in some places. While replacing history information with state information is tempting to exploit additional information during centralized training, it has been shown to be unsound in partially observable settings, as recently shown in the actor-critic CTDE case \citep{stateissues}.   

To address the gap between theory and practice in value factorization, we extend the theory to the \emph{stateful} case that combines state and history information. We show that IGM still holds for most of the methods (VDN, QMIX, and QPLEX) but not necessarily for the weighted version of QMIX (WQMIX) \citep{wqmix}. We also show that QPLEX's practical implementation can not represent the full IGM function class due to the use of state information instead of history information---losing one of its main benefits. 

While there are many architectures that would satisfy IGM, previous approaches, such as QPLEX, made choices based on earlier work without exploring other alternatives that could improve performance. 
For example, unlike dueling networks, which typically learn separate history and advantage value functions at the agent level, QPLEX learns a Q-function and assumes the V-function is a max of it. 
For this reason, we introduce DuelMIX. DuelMIX maintains separate estimators at the agent level---instead of computing them from the agents' Q-functions. Such a separation has been shown to learn better value approximations, which enhance performance and sample efficiency in single-agent scenarios \citep{dueling}. 

Our work makes the following contributions: we (i) formalize IGM over the centralized stateful functions used in practical implementations; (ii) analyze that the state does not introduce bias into QMIX \citep{qmix} for IGM, and QPLEX \citep{qplex} for Advantage-IGM; (iii) empirically show that using other sources of information during factorization (i.e., constant and random vectors) could lead to performances comparable or better than using the state, contrasting the common belief that the state allows higher performance; (iv) present DuelMIX, a factorization scheme integrating dueling networks at a per-agent level, and combining joint history-state values in a weighted fashion to achieve full expressiveness. 
This learning stream separation leads to significant benefits in cooperative scenarios where optimal joint policies often hinge on specific actions.

To validate our intuitions, we test the methods in the highly partially observable Box Pushing (BP) scenario, where the optimal behavior is contingent on a specific agent's action \citep{yuchen_ma}, and micromanagement tasks in the recent StarCraft II Lite (SMACLite) suite \citep{smaclite}. Our results on BP show the benefits of separate value learning, allowing DuelMIX to achieve good performance where previous approaches fail. Moreover, SMACLite experiments show that DuelMIX outperforms previous factorization methods and significantly improves sample efficiency.

%% file: Sections/preliminaries.tex
\section{Preliminaries}
\label{sec:preliminaries}

We model fully cooperative multi-agent tasks as \textit{Decentralized Partially Observable Markov Decision Processes} (Dec-POMDPs) \citep{decpomdp}, represented as tuple $\langle \mathcal{N}, \mathcal{S}, \mathcal{U}, T_\mathcal{S}, r, \mathcal{O}, T_\mathcal{O}, \gamma \rangle$. $\mathcal{N}$ is a finite set of agents; $\mathcal{S}$ is a finite set of states; $\mathcal{U} \equiv \langle U_i \rangle_{i \in \mathcal{N}}$ and $\mathcal{O} \equiv \langle O_i \rangle_{i \in \mathcal{N}}$ are the finite sets of joint actions and observations respectively, while $U_i, O_i$ are the individual ones. At each time step, every agent $i$ chooses an action, forming the joint action $\bm{u} \equiv \langle u_i \rangle_{i \in \mathcal{N}} \in \mathcal{U}$. After performing $\bm{u}$, the environment transitions from a state $s$ to a new $s'$, following a transition probability function $T_\mathcal{S} \colon \mathcal{S} \times \mathcal{U} \times \mathcal{S} \rightarrow [0, 1]$ ($T_\mathcal{S}(s, \bm{u}, s') = \Pr (s'\mid s, \bm{u})$), and returning a joint reward $r\colon \mathcal{S} \times \mathcal{U} \rightarrow \mathbb{R}$ for being in state $s \in \mathcal{S}$ and taking actions $\bm{u} \in \mathcal{U}$. In a partially-observable setting, agents receive an observation $\bm{o} \equiv \langle o_i \rangle_{i\in\mathcal{N}} \in \mathcal{O}$ according to an observation probability function $T_\mathcal{O}\colon \mathcal{S} \times \mathcal{U} \times \mathcal{O} \rightarrow [0, 1]$ ($T_\mathcal{O}(\bm{u}, s', \bm{o}) = \Pr(\bm{o}\mid \bm{u}, s')$), and each agent maintains a policy $\pi_i(u_i\mid h_i)$ mapping local action-observation histories $h_i = (o_{i,0}, u_{i,0}, o_{i,1}, \ldots, o_{i,t}) \in H_i$ to actions. In finite-horizon tasks, we aim to find a joint policy $\pi(\bm{u}\mid \bm{h})$ maximizing the expected discounted episodic return $\mathbb{E}_{\pi}\left[\sum_t \gamma^t r_t \right]$, where $\gamma \in [0, 1)$ is the discount and $\bm{h}_t = \langle 
\bm{o}_0, \bm{u}_0, \dots, \bm{o}_t \rangle \in \mathcal{H}$ is the joint action-observation history.

\subsection{Value Factorization}
\label{sec:background_vf}

In this section, we summarize the stateless theoretical framework presented by seminal value factorization works. In \Cref{sec:methods_analysis}, we will discuss how most algorithms use the state during factorization. As such, we argue the following preliminaries do not correctly reflect most of the published literature.

Factorization algorithms must satisfy the IGM principle (\Cref{eq:igm}) \citep{qtran}, ensuring consistency between decentralized and centralized decision-making.
\begin{equation}
    \argmax_{\bm{u} \in \mathcal{U}}Q(\bm{h}, \bm{u}) \equiv 
    \left(
    \argmax\limits_{u_i \in U_i}Q_i(h_i, u_i)   
    \right)_{i\in\mathcal{N}}.
\label{eq:igm}
\end{equation}
This consistency is pivotal for scalability as it facilitates tractable joint action selection by deriving the joint greedy action from each agent's local utility. The individual history-action utilities 
$\langle Q_i:H_i\times U_i \rightarrow \mathbb{R}\rangle_{i\in\mathcal{N}}$ satisfy IGM for a joint history-action value function $Q:\mathcal{H}\times\mathcal{U}\rightarrow \mathbb{R}$ if the maximal actions over the centralized function and the local utilities align. Several methods have been proposed, each imposing different architectural constraints to guarantee IGM during the centralized training process. We provide concise descriptions of the main approaches considered by our work.

\textbf{Additive Constraint.} VDN is the foundational factorization method and expresses the joint history-action value as a sum of per-agent utilities \citep{vdn}:
    \begin{equation}
        Q(\bm{h}, \bm{u}) = \sum_{i=1}^{|\mathcal{N}|} Q_{i}(h_i, u_i).
    \label{vdn}
    \end{equation}

We note that VDN is correctly formalized in terms of stateless functions and utilities, as it does not employ state information. However, VDN only represents a limited set of joint functions well. 

\textbf{Monotonic Constraint.} QMIX uses a non-linear monotonic mixing network to combine agent utilities \citep{qmix}:
\begin{equation}
    \frac{\partial Q(\bm{h}, \bm{u})}{\partial Q_i(h_i, u_i)} \geq 0, \forall i \in \mathcal{N}.
\label{eq:monotonicity}
\end{equation}
By enforcing positive weights to satisfy the constraint in \Cref{eq:monotonicity}, QMIX represents a broader class of functions compared to VDN. However, it is limited to functions that can be factored as non-linear monotonic combinations of the agents’ utilities. Recently, Weighted-QMIX (WQMIX) extended QMIX with a weighting mechanism, placing more importance on better joint actions \citep{wqmix}.
Despite framing the monotonic factored value in a stateless fashion, these methods use the state in the mixing network. 
This prompts our investigation of whether the state introduces possible learning biases, discussed later in \Cref{sec:methods_analysis}.

\textbf{Advantage-IGM.} An advantage-based IGM principle equivalent to \Cref{eq:igm} has been proposed by \citet{qplex}. 

Given individual utilities $\langle Q_i \rangle_{i\in\mathcal{N}}$ and the joint $Q$ defined as:
\begin{align}
    Q(\bm{h}, \bm{u}) &= V(\bm{h}) + A(\bm{h}, \bm{u}), &
    Q_i(h_i, u_i) &= V_i(h_i) + A_i(h_i, u_i),
    \label{eq:q_decomp}
\end{align}
with $V, A$ not strictly representing the mathematical values we typically associate with them, but their definition regardless of the policy:
\begin{equation}
\begin{aligned}
    V(\bm{h}) &= \max_{\bm{u}'}Q(\bm{h}, \bm{u'}), &
    A(\bm{h}, \bm{u}) &= Q(\bm{h}, \bm{u}) - \max_{\bm{u}'}Q(\bm{h}, \bm{u'}), \\
    V_i(h_i) &= \max_{u_i'}Q_i(h_i, u_i'), &
    A_i(h_i, u_i) &= Q_i(h_i, u_i) - \max_{u_i'}Q_i(h_i, u_i').
\end{aligned}
\label{eq:values_def}
\end{equation}
Advantage-IGM is satisfied if the equivalence between centralized and decentralized action selections with respect to advantage functions holds:
\begin{equation}
    \argmax_{\bm{u} \in \mathcal{U}} ~A(\bm{h}, \bm{u}) \equiv 
    \left(
    \argmax\limits_{u_i \in U_i}A_i(h_i, u_i) 
    \right)_{i\in\mathcal{N}}.
    \label{eq:adv_igm}
\end{equation}
In its original formalization using \Cref{eq:values_def}, \Cref{eq:adv_igm} has been shown equivalent to \Cref{eq:igm} when the advantage values are non-positive \citep{qplex}---optimal actions' advantage must be zero, and non-optimal actions must have negative advantages. 

QPLEX \citep{qplex} builds on Advantage-IGM and decomposes the per-agent utilities $Q_i(h_i, u_i)$ into individual $V_i(h_i), A_i(h_i, u_i)$ according to \Cref{eq:values_def}. In the practical implementation, such values are then conditioned on joint information using a non-linear transformation. This module outputs biases and positive weights $\langle b_i(\bm{h}), w_i(\bm{h}) > 0 \rangle_{i\in\mathcal{N}}$:
\begin{equation}
    V_i(\bm{h}) = w_i(\bm{h})V_i(h_i)+b_i(\bm{h}), \quad
    A_i(\bm{h}, u_i) = w_i(\bm{h})A_i(h_i,u_i). 
\label{eq:qplex_transf}
\end{equation}

After this transformation, QPLEX factorizes the joint history-action value function as:
\begin{equation}
    Q(\bm{h}, \bm{u}) = \sum_i V_i(\bm{h}) + \lambda_i(\bm{h}, \bm{u})A_i(\bm{h}, u_i),
\label{eq:qplex_joint}
\end{equation}
where $\langle \lambda_i(\bm{h}, \bm{u}) > 0 \rangle_{i\in\mathcal{N}}$ are weights computed with an attention module to enhance credit assignment.\footnote{Both the transformation and the attention module use positive weights to maintain action selection consistency over the advantage values. Naturally, this positivity enforces monotonicity in the advantage factorization.} While QPLEX achieves higher empirical performance than previous factorization methods, the authors claimed the transformations and $\lambda_i(\bm{h}, \bm{u})$ coefficients allow QPLEX to fully represent the functions satisfying Advantage-IGM \citep{qatten}. This would be correct when using the joint history in the factorization modules, but the actual implementation uses the state. For this reason, in \Cref{sec:methods_analysis}, we argue the QPLEX practical implementations are incorrect \citep{qplex_code1, qplex_code2, qplex_code3, qplex_code4} since they estimate weights and biases as functions of the state and not of the joint history (i.e., in practice we have $\langle b_i(s), w_i(s) > 0, \lambda_i(s, \bm{u}) \rangle_{i\in\mathcal{N}}$ rather than $\langle b_i(\bm{h}), w_i(\bm{h}) > 0, \lambda_i(\bm{h}, \bm{u}) \rangle_{i\in\mathcal{N}}$).
Along these lines, QPLEX's theoretical framework also uses stateless functions, diverging from the actual stateful implementation. To address the theoretical mismatch, we formalize the stateful Advantage-IGM in \Cref{sec:methods_analysis}, and formally analyze whether the state introduces biases also for QPLEX \citep{qplex}.

Additionally, we believe that QPLEX's performance can be further improved since: (i) the left side of \Cref{eq:qplex_joint} is factored in a simple additive fashion, and (ii) history and advantage utilities are decomposed from every $Q_i(h_i, u_i)$ by their definition. These values are not learned locally and separately by each agent as originally proposed by dueling networks \citep{dueling}. 

A true dueling architecture is relevant when a small subset of actions are useful for the task, which we note commonly happens in highly cooperative scenarios. To address these issues, we introduce a novel dueling networks-based factorization architecture in \Cref{sec:methods_duelmix}.

%% file: Sections/methods.tex
\section{Stateful Value Factorization}
\label{sec:methods_analysis}

Practical implementations of QMIX, WQMIX, and QPLEX often use state values in their centralized models---usually through a mixing network---in ways that are neglected in their theoretical analysis (see \Cref{sec:methods} for a formal summary of stateless and stateful QMIX and QPLEX variants). For the first time (to our knowledge), we analyze if the theory of these methods extends correctly to the stateful case, or whether using state values introduces any learning bias. Such biases can easily happen by improper uses of state in single and multi-agent partially observable control problems as recently shown by \citet{baisero_unbiased_2022, baisero_asymmetric_2022, stateissues}.

We begin by adjusting the notation of centralized value models that use state, effectively resulting in history-state values $Q(\bm{h}, s, \bm{u})$. Given that consistent history and history-state values are related by $Q(\bm{h}, \bm{u}) = \mathbb{E}_{s\mid \bm{h}}\left[ Q(\bm{h}, s, \bm{u}) \right]$, we correctly reformulate the IGM principle in a marginalized history-state form as follows:

\begin{proposition}[History-State IGM]
    For a joint $Q:\mathcal{H}\times \mathcal{S}\times \mathcal{U} \rightarrow \mathbb{R}$ and individuals $\langle Q_i:H_i\times U_i\rightarrow \mathbb{R}\rangle_{i\in\mathcal{N}}$ such that the following holds:
    \begin{equation}
        \argmax_{\bm{u} \in \mathcal{U}} \mathbb{E}_{s\mid \bm{h}}\left[ Q(\bm{h}, s, \bm{u}) \right] \equiv 
        \left(
        \argmax\limits_{u_i \in U_i}Q_i(h_i, u_i)
        \right)_{i\in\mathcal{N}},
    \label{eq:igm:Ehs}
    \end{equation}
    $\langle Q_i(h_i, u_i)\rangle_{i\in\mathcal{N}}$ are said to satisfy History-State IGM for $Q(\bm{h}, s, \bm{u})$.
\label{prop:state_igm}
\end{proposition}
\vspace{-5pt}

\textbf{Role of the State in QMIX.} The question now becomes whether the mixing model of QMIX (see \Cref{sec:state-qmix} for details) satisfies the History-State IGM principle. Given QMIX's hyper-network architecture and the way the state is used, the monotonicity constraint of \Cref{eq:monotonicity} holds for every state even for history-state values, resulting in a \emph{non-marginalized} version of History-State IGM:
\begin{equation}
    \argmax_{\bm{u} \in \mathcal{U}} Q(\bm{h}, s, \bm{u}) \equiv 
    \left(
    \argmax\limits_{u_i \in U_i}Q_i(h_i, u_i)
    \right)_{i\in\mathcal{N}}.
\label{eq:igm:hs}
\end{equation}

Notably, \Cref{eq:igm:hs} is distinct from \Cref{eq:igm:Ehs} in ways that may have problematic or even catastrophic consequences in partially observable control, e.g., the maximal action of a centralized history-state value corresponds to the optimal joint action under full observability, and not necessarily the optimal joint action for a team of partially observable agents.  Fortuitously, despite the misleading formalization of the stateless IGM in the QMIX work \citep{qmix}, we confirm that the architectural constraints of QMIX's architecture restrict the centralized model in such ways that fundamentally avoid this issue altogether; a formal proof is provided in \Cref{sec:methods_analysis:proof}.
\begin{proposition}[QMIX State Bias]
\label{thm:qmix-state-bias}
A stateful implementation of QMIX does not introduce any additional state-induced bias compared to a stateless implementation.
\end{proposition}
\textbf{Role of the State in WQMIX.} WQMIX~\citep{wqmix} is an extension of QMIX that generalizes the representation space of QMIX models by applying a loss weighting scheme. The authors provide a theoretical analysis that proves correctness, unbiasedness, and the ability of their centralized models to act as universal function approximators. Their analysis is already framed in the context of stateful values that are relevant to our work. However, it is also limited by the further assumption that the agents have full observability of the state to begin with; a significant discrepancy compared to their proposed methods in practice. To the best of our knowledge, there is yet no extension of their analysis that holds without the full observability assumption.  Further, the methods that we have used to prove the unbiasedness of both QMIX and, following, QPLEX do not similarly hold for WQMIX.

\textbf{Role of the State in QPLEX.} The analysis for QPLEX follows a similar structure to that for QMIX. We begin by adjusting the model notation by replacing the joint history in the weights and biases estimated by the transformation and mixing modules with the state (see \Cref{sec:state-qplex} for details) as used in the actual implementations \cite{qplex_code1, qplex_code2, qplex_code3}. The accurate stateful formulations of \Cref{eq:qplex_transf,eq:qplex_joint} are the following:
\begin{align}
    \label{eq:qplex_transf:hs}
    V_i(h_i, s) &= w_i(s)V_i(h_i)+b_i(s), \quad A_i(h_i, s, u_i) = w_i(s)A_i(h_i,u_i), \\
    \label{eq:qplex_joint:hs}
    Q(\bm{h}, s, \bm{u}) &= \sum_i V_i(h_i, s) + \lambda_i(s, \bm{u})A_i(h_i, s, u_i).
\end{align}
We then argue that these state-based implementations do not result in full IGM expressiveness. The advantage stream of QPLEX is composed by a monotonic combination of individual state-history advantage utilities (with the same expressiveness of monotonic factorization) and an additive combination of local state-history utilities. One of the two streams (i.e., $V$ or $A$) would require feeding joint history information in an unconstrained non-linear mixer to achieve full expressiveness.

Regarding the potential learning bias introduced by using the state, the architectural constraints of QPLEX have to guarantee the History-State Advantage-IGM principle, which is again distinct from the respective stateless formulation of \Cref{eq:adv_igm} in ways that may negatively impact partially observable control. 

Nonetheless, as in the case of QMIX, we can show that the QPLEX models prohibit the state from informing the action-selection process; a formal proof is provided in \Cref{sec:methods_analysis:proof}.
\begin{proposition}[QPLEX State Bias]
\label{thm:qplex-state-bias}
A stateful implementation of QPLEX does not introduce any additional state-induced bias compared to a stateless implementation.
\end{proposition}

\section{DuelMIX}
\label{sec:methods_duelmix}
To overcome the implementation drawbacks of previous factorization algorithms, we present a novel factorization scheme, DuelMIX. Our approach leverages the dueling networks estimator at a per-agent level and introduces a weighted mixing mechanism that estimates a joint history value. Following our intuitions on stateful factorization, we introduce DuelMIX using stateful functions.

\begin{figure}[b]
    \vspace{-5pt}
    \centering
    \includegraphics[width=.8\linewidth]{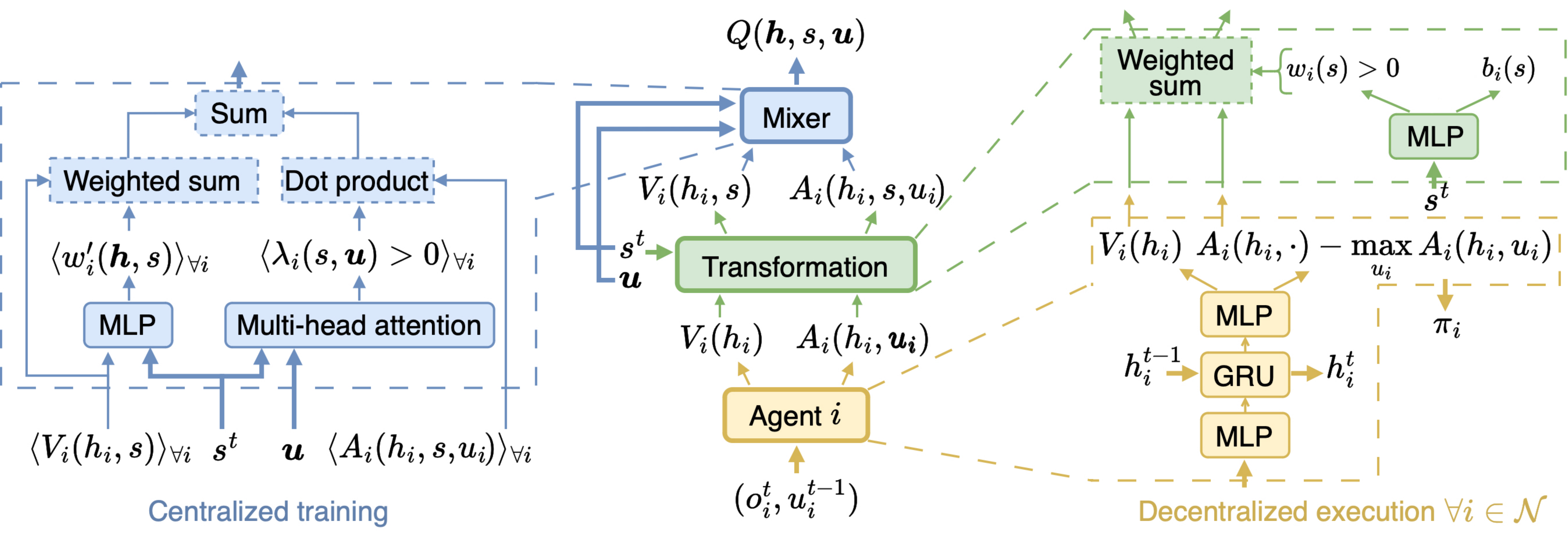}
    \caption{DuelMIX architecture: (i) agent dueling utility network structure (yellow); (ii) transformation module (green); (iii) mixing network architecture.}
    \label{fig:duelmix_p1}
    \vspace{-3pt}
\end{figure}

\textbf{Algorithm.} The overall architecture of DuelMIX is detailed in the following sections. In particular, \Cref{fig:duelmix_p1} shows the overall architecture of DuelMIX, which is composed of the following modules:

\textbf{Agent Dueling Utility} (yellow). Each agent $i\in\mathcal{N}$ employs a recurrent $Q$-network taking its previous hidden state $h_i^{t-1}$, previous action $u_i^{t-1}$, and current observation $o_i^t$ as input to ensure decentralized execution. In contrast to previous factorization methods that produce a local utility $Q_i(h_i, u_i)$ through a single-stream estimator, DuelMIX utilizes two separate streams and outputs history and advantage utilities, denoted as $V_i(h_i)$ and $A_i(h_i, u_i)$, respectively.
Crucially, building upon the insights of \citet{dueling}, each centralized update influences the $V_i$ stream, enhancing the approximation of the history value. In single-agent scenarios, enhancing the approximation of value functions has led to higher performance and sample efficiency \citep{dueling, marche23evopg, marzari24retrain}, and also proves to be particularly advantageous in cooperative tasks. In these tasks, the optimal joint policy often hinges on specific actions taken in particular histories (as demonstrated in the Box Pushing experiments of \Cref{sec:experiments}). Simultaneously, advantage utilities are employed for decentralized action selection, such as with an $\epsilon$-greedy policy. In particular, the advantage stream outputs $A_i(h_i,\cdot) - \max_{u_i} A_i(h_i,u_i)$, forcing advantages to be zero for the chosen action and non-positive ($\leq 0$) for the others (which becomes necessary when transforming the advantage values with positive weights).

\textbf{Transformation} (green). DuelMIX incorporates the transformation network used by Qatten and QPLEX. 
In detail, the transformation network combines local utilities $\langle V_i(h_i), A_i(h_i, u_i)\rangle_{i\in\mathcal{N}}$ with state information $\langle V_i(h_i, s), A_i(h_i, s, \bm{u})\rangle_{i\in\mathcal{N}}$. This module consists of a multi-layer perceptron (MLP) taking the state $s$ as input, and outputs a set of biases and positive weights $\langle b_i(s), w_i(s) > 0 \rangle_{i\in\mathcal{N}}$. These weights and biases transform the local utilities as in \Cref{eq:qplex_transf:hs}. Unlike QPLEX, the DuelMIX transformation module transforms the $V_i(h_i), A_i(h_i, u_i)$ learned by the agent, instead of being obtained by decomposition from $Q_i(h_i, u_i)$ following their optimal definition as in \Cref{eq:q_decomp}. 

\textbf{Mixing} (blue). The centralized mixing network employed by DuelMIX uses the same efficient multi-head attention mechanism as QPLEX to estimate positive importance weights $\langle \lambda_i(s, \bm{u}) > 0  \rangle_{i\in\mathcal{N}}$. Maintaining positivity is essential for action-selection consistency. These weights, combined with the per-agent transformed advantages, yield the stateful joint advantage value:
\begin{equation}
    A(\bm{h}, s, \bm{u}) = \sum_{i}^{|\mathcal{N}|}\lambda_i(s, \bm{u})A_i(h_i, s, u_i).
\label{eq:duelmix_joint_a}
\end{equation}
In contrast to prior works, we use an MLP taking as input the transformed history utilities and the state to estimate weights $\langle w_i'(\bm{h}, s)\rangle_{i\in\mathcal{N}}$ that are used to factorize the joint history value as:
\begin{equation}
    V(\bm{h}, s) = \sum_{i}^{|\mathcal{N}|}w_i'(\bm{h},s)V_i(h_i, s).
\label{eq:duelmix_joint_v}
\end{equation}
Unlike the positive weights in the joint advantage factorization, our design of $w_i'(\bm{h},s)$ is driven by two key motivations:\footnote{We explored different variations in designing the centralized history value stream of DuelMIX, with \Cref{eq:duelmix_joint_v} yielding the best overall performance.}  they (i) can assume arbitrary values since $V(\bm{h}, s)$ does not influence the action-selection process; (ii) allow DuelMIX to have full expressiveness over the class of functions satisfying IGM since they are a function of the joint history $\bm{h}$ and are not limited by monotonic constraints. Hence, this value stream can correct for the discrepancies between the centralized joint action value function and the monotonic advantage combination.

The joint history-state-action value driving the centralized learning process then follows by definition:
\begin{equation}
    Q(\bm{h}, s, \bm{u}) = V(\bm{h}, s) + A(\bm{h}, s, \bm{u}).
\label{eq:duelmix_jointq}
\end{equation}

Given the nature of factorization methods based on Deep Q-Networks \citep{dqn}, DuelMIX is trained end-to-end to minimize the mean squared error loss:
\begin{equation}
    L(\Theta) = \frac{1}{|b|}\sum_{i=1}^{|b|}\left[\left(y-Q_{\Theta}(\bm{h}, s, \bm{u})\right)^2 \right], \quad y = r + \gamma Q_{\Theta'}(\bm{h}', s', \bm{u}'),
\end{equation}
where ${\bm{u}'=\langle u_i' = \argmax_{u_i'}Q_i(h_i', u_i')\rangle_{i\in\mathcal{N}}}$, $\Theta$ represents the weights of the entire DuelMIX network ($\Theta'$ are the parameters of a target network \citep{ddqn}), and $b$ is a batch of transitions sampled from a replay buffer. Given the nature of our architecture, DuelMIX shares the same limitations as related value factorization works (e.g., having a simulator for centralized training, where it is possible to get the state of the environment and communicate between agents).

\subsection{Representational Complexity}
\label{sec:representational_complexity}
The positive weights in DuelMIX, the non-positive advantage utilities, and the joint history used to estimate $w_i'(\bm{h}, s)$, enable our method to satisfy IGM and achieve full expressiveness. A formal proof is provided in \Cref{sec:duelmix_proof}.
\begin{proposition}
    The function class that DuelMIX can realize is equivalent to what is induced by History-State Advantage-IGM (\Cref{eq:app:hs:aigm}).
\label{prop:duelmix_expressiveness}
\end{proposition}

\textbf{Role of the State in DuelMIX.} We conclude our analysis of DuelMIX by including a result analogous to those in \Cref{sec:methods_analysis}, on the bias that may be introduced by a potentially improper use of stateful models.
As in the case of both QMIX and QPLEX, we are able to determine the following proposition, whose formal proof is provided in \Cref{sec:methods_analysis:proof}.
\begin{proposition}[DuelMIX State Bias]
\label{thm:duelmix-state-bias}
A stateful implementation of DuelMIX does not introduce any additional state-induced bias compared to a stateless implementation.
\end{proposition}

%% file: Sections/experiments.tex
\section{Experiments}
\label{sec:experiments}
This section presents a comprehensive evaluation of the performance of DuelMIX in comparison to existing factorization methods, namely VDN \citep{vdn}, QMIX \citep{qmix},\footnote{We use QMIX over WQMIX as it achieved comparable performance when fine-tuned while being less computationally demanding \citep{qmix_finetune}.} Qatten \citep{qatten}, and QPLEX \citep{qplex}.

Our experiments address the following key questions: (i) \textit{Does the agent's dueling utility network learn a more effective representation of the state value?} (ii) \textit{Does DuelMIX exhibit better performance over previous algorithms?} (iii) \textit{Is the state crucial for performance in value factorization?} To answer these questions, we conduct experiments using the well-known Box Pushing task \citep{boxpushing} and standard micromanagement tasks based on Starcraft II \citep{smaclite}.

\textbf{Implementation Details.}
Data collection is performed on Xeon E5-2650 CPU nodes with 64GB of RAM, using existing implementations for the baselines \citep{vdn, qmix, qplex, qatten}. Hyperparameters are in \Cref{suppl:hyperparameters} and we report the average return smoothed over the last ten episodes of ten runs per method. Shaded regions represent the standard error. This number of independent trials surpasses the typical 3-5 runs used in previous works \citep{vdn, qmix, qplex, qatten}. Considering the computational resources used for our experiments, \Cref{suppl:env_impact} addresses the environmental impact and our strategy to offset estimated CO2 emissions.

\subsection{Environments}
To demonstrate the benefits of learning per-agent separate utilities, we consider the BP task \citep{boxpushing} with a grid size of 30 (BP-30). In this Dec-POMDP task, two agents must collaborate to move a large box to the goal. Notably, agents can individually push small boxes, while moving the large box requires synchronized effort. Agents have very limited visibility observing only the cell in front of them, making high-dimensional scenarios considerably challenging. 
We also test in the SMACLite decentralized micromanagement tasks \citep{smaclite}. SMAC is the standard benchmark for evaluating factorization-based MARL algorithms and SMACLite significantly reduces computational requirements while maintaining comparable performance to the original version \citep{starcraft}.\footnote{SMACLite trained policies have comparable results in the same SMAC scenarios \citep{smaclite}.} We consider seven different setups, including two unsolved super-hard tasks proposed by \citet{qplex} to show the superior performance of DuelMIX. We refer to \Cref{sec:environments} for a more detailed discussion of these scenarios.

\subsection{Results}

\begin{figure}[t]
    \centering
    \begin{minipage}{0.3\linewidth}
        \centering
\includegraphics[width=\linewidth]{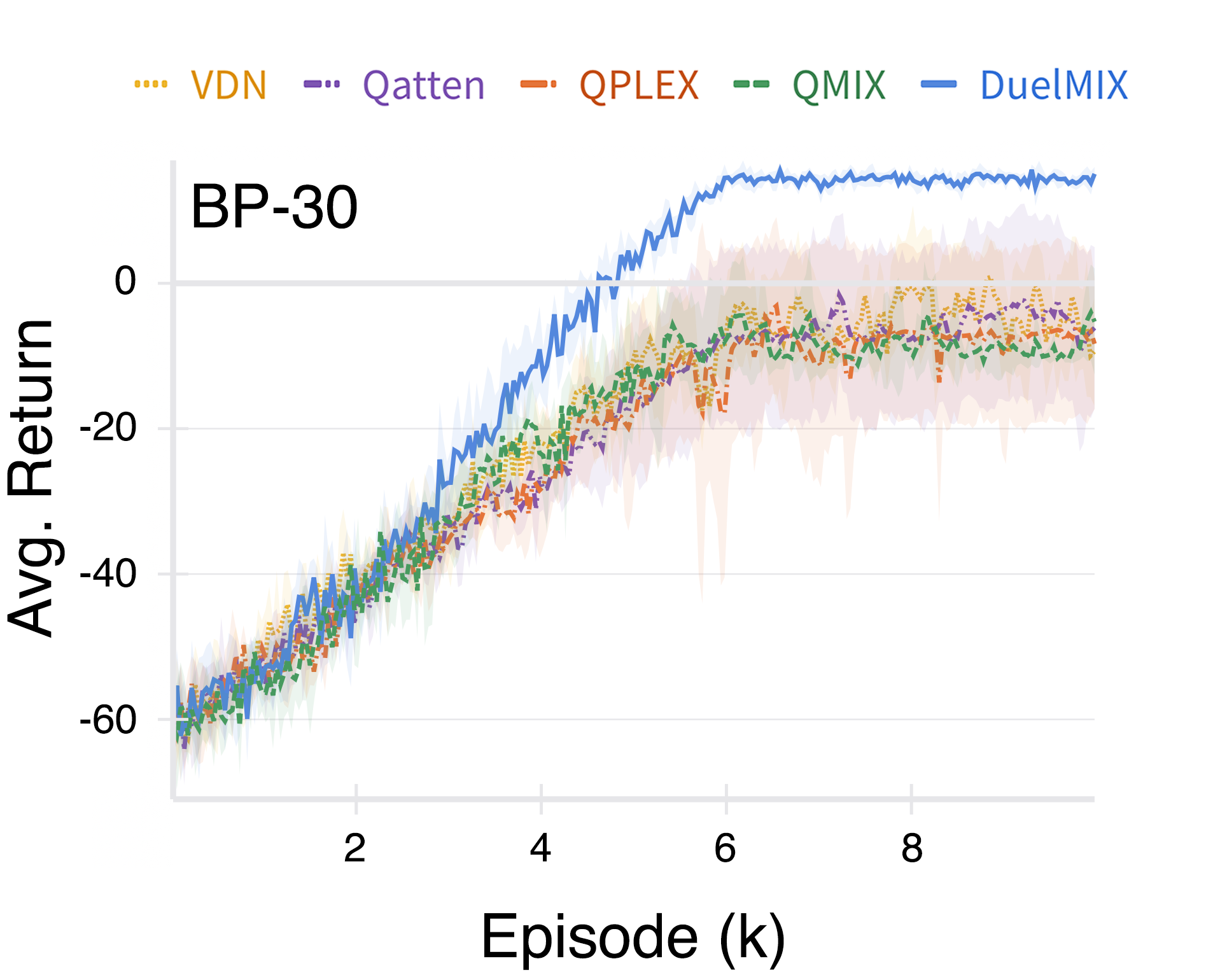}
        \caption{Learning curves for the stateful algorithms in BP.}
        \label{fig:res_bp30}
    \end{minipage}
    \hspace{5pt}
    \begin{minipage}{0.45\linewidth}
        \centering
\includegraphics[width=\linewidth]{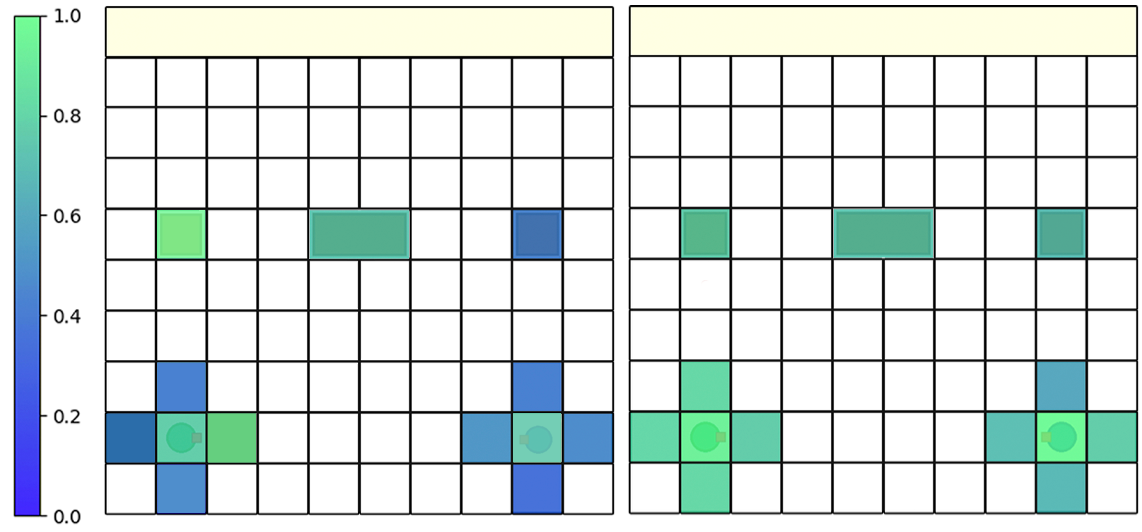}
        \caption{Saliency map of DuelMIX (left) and QPLEX (right) left agent's state value with respect to the initial state.}
        \label{fig:bp_saliency}
    \end{minipage}
\vspace{-5pt}
\end{figure}

\textbf{Box Pushing.}
This challenging Dec-POMDP task allows us to visualize how much importance the dueling utility network gives to the input features.
\Cref{fig:res_bp30} shows results for stateful factorization algorithms. Overall, previous methods learn very sub-optimal policies, where both agents roam around in the grid for a few steps, before one of them successfully pushes a small box to the goal. DuelMIX learns a more effective policy that enables the two agents to navigate directly to small boxes and push them simultaneously to the goal. This confirms DuelMIX's capability to achieve higher returns in challenging scenarios.
\begin{figure*}[b]
    \vspace{-6pt}
    \centering
    \includegraphics[width=0.95\linewidth]{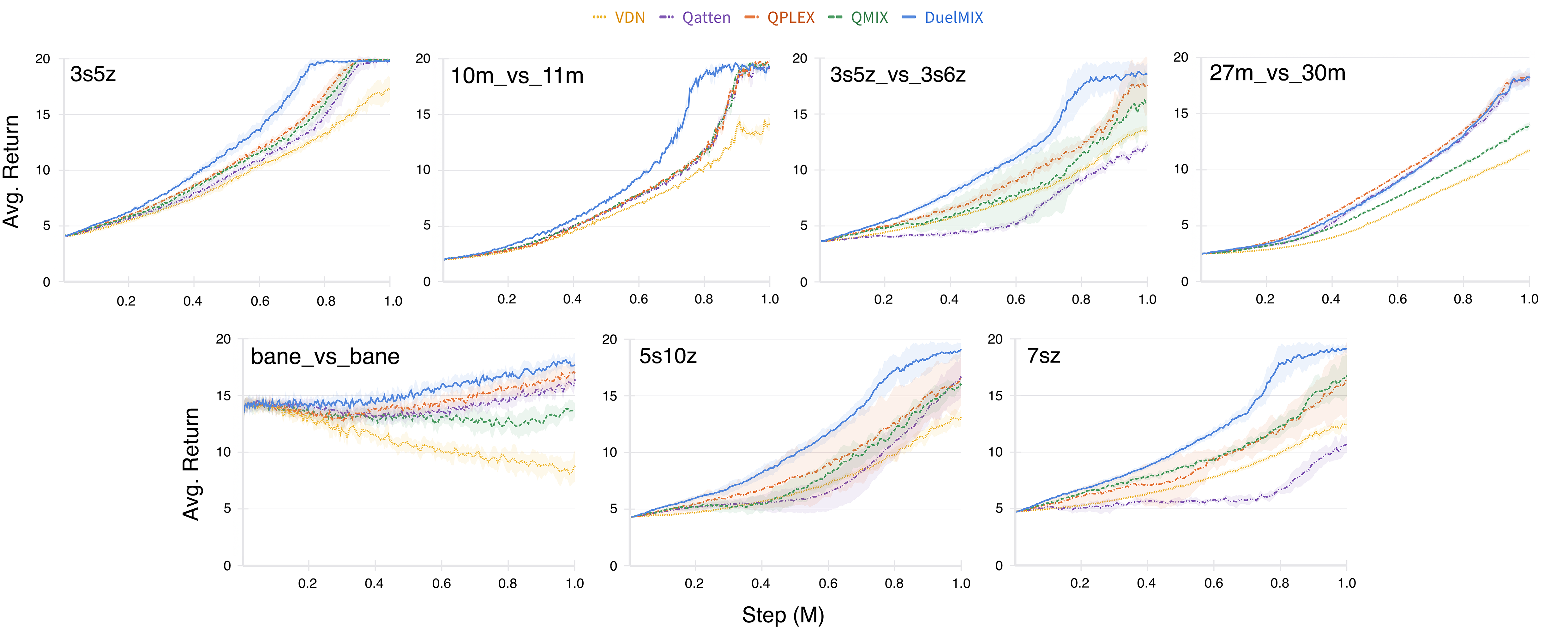}
    \vspace{-0.3cm}
    \caption{Average return during training for stateful factorization algorithms in SMACLite maps.}
    \label{fig:res_state}
    \vspace{-6pt}
\end{figure*}

\textbf{State Stream Representation.}
We show the saliency maps of \citet{saliency}, employing the Jacobian of the trained network, which allows us to visualize salient parts of the input as seen by the model’s weights, showing which input features are more relevant for the weights.
For clarity, this requires re-training our policies in a fully observable setup, where agents take as input their positions, their orientations as a one-hot encoding, and the position of the boxes.\footnote{Previous runs only observe the cell in front of the agents, which does not provide useful visual information.} It is thus possible to plot the ``importance" that the state value stream of DuelMIX and QPLEX gives at each input. The idea is to show that DuelMIX's state value stream gives significantly less importance to entities that are not relevant to achieving a good payoff, while QPLEX's state value is not able to do so.
\Cref{fig:bp_saliency} overlaps the saliency map onto a reduced-size BP view. Blue cells indicate the features (cells) are not relevant to the agent so they will not affect its decisions relevantly. In contrast, green cells indicate the agent gives a high importance to those features. So we would like to see our policies giving more importance (i.e., green) to the only cells leading to a positive reward.
We show the average normalized importance that DuelMIX’s value (left) and QPLEX's one (right) give to the input features at the first step in the environment for the agent on the left. Crucially, we obtained similar results at different steps, showing DuelMIX’s state value stream gives significantly higher importance to features that are relevant for achieving high returns (high-valued features in green are agents' positions and the nearby boxes while the others have inferior value and marked in blue). In contrast, QPLEX almost equally balances the importance of all the input features (i.e., all the values are similar), which could be detrimental to training.

\textbf{SMACLite.}
\Cref{fig:res_state} illustrates the results of our evaluation in SMACLite environments. Notably, DuelMIX achieves the highest average return across most tasks, demonstrating superior performance. Moreover, DuelMIX exhibits improved sample efficiency, especially in easy and hard environments, learning behaviors with higher payoffs in fewer steps. 
In contrast, the limited expressiveness of VDN is known to struggle in complex domains. Qatten also achieved low performance. This is potentially related to the discrepancy between the centralized joint history-state-action value function and the Qatten formalization, which has been addressed by QPLEX. The latter achieved the highest overall performance among the factorization baselines, due to the benefits of building upon Advantage-IGM. Moreover, it is interesting to note QMIX's competitive performance over the more advanced QPLEX. These results further confirm the intuitions of \citet{qmix_finetune}, which showed QMIX significantly benefits from appropriate fine-tuning.

\subsection{Influence of the State on Performance}
\label{sec:stateperformance}
We explore the performance of using different centralized information during factorization. Injecting the state in the mixer has become a standard de facto, but its use can not be supported by the theory.
We investigate the performance of factorization algorithms (except VDN, which does not use the state) on representative SMACLite tasks. Following our stateful analysis in \Cref{sec:methods_analysis}, the state information gets marginalized out so it is not clear that any information is actually being used from it. In both scenarios, the impact of weights and biases outputted by the factorization modules introduce noise in the joint estimation with the result of increasing exploration and, potentially, robustness. As such, we expect different sources of centralized information to work well with factorization algorithms.

Table \ref{tab:res_stateinfo} in Appendix \ref{sec:state_results} presents the results, employing two types of centralized information in place of the state ($s$): (i) uniform random noise $\in [0, 1.0]$ ($r$), and (ii) a constant zero vector ($c$). 
In particular, uniform random noise has maximum entropy, while the zero vector carries no information (generating weights and biases depending on the learned ``internal" biases).
These experiments use the same hyperparameters of the previous evaluation. Interestingly, when using random noise, different methods result in different behaviors. QMIX is affected detrimentally, and its performance is significantly lower than its stateful version. The performance of Qatten remains similar in most scenarios. QPLEX achieves comparable performance in \textit{5s10z} while being affected slightly negatively in the other two tasks. While behaving similarly to QPLEX in the latter tasks, DuelMIX with random noise achieves higher performance in \textit{5s10z}. Moreover, the performance gap between the zero and the random cases is negligible for DuelMIX, which achieves superior performance in \textit{5s10z}. The only major difference is that QPLEX outperforms its stateful version in the same task.
\begin{wraptable}{r}{0.35\textwidth}
        \vspace{-5pt}
      \centering
        \caption{Average return for fine-tuned QPLEX and QMIX with (r, c) vector information.}
        \label{tab:res_finetune}
        \begin{tabular}{lll}
        \toprule
               \textbf{(fine-tuned$~\downarrow$)}  & \multicolumn{1}{c}{\textbf{}}  & \textbf{5s10z}                     \\ \midrule
        \textbf{QMIX}  & \multicolumn{1}{c}{\textit{s}} & \multicolumn{1}{c}{\textbf{15.8 $\pm$ 0.4}} \\
        \textbf{}                   & \textit{r}                     & 14.5 $\pm$ 1.4                     \\
                                    & \textit{c}                     & 14.7 $\pm$ 0.1                     \\ \midrule
        \textbf{QPLEX} & \textit{s}                     & 16.2 $\pm$ 2.1                     \\
                                    & \textit{r}                     & 18.0 $\pm$ 0.6                     \\
                                    & \textit{c}                     & \textbf{18.3 $\pm$ 0.8}                     \\ \bottomrule
        \end{tabular}
    \vspace{-5pt}
\end{wraptable}

\textbf{Fine-tuning.} Additional fine-tuning experiments in Table \ref{tab:res_finetune} reveal that using these different centralized information result in comparable or superior performance over the stateful choice, emphasizing the importance of our empirical investigation. We performed our initial grid search to fine-tune random noise and zero vector-based factorization methods. \Cref{tab:res_finetune} shows the positive outcomes of these experiments in \textit{5s10z}. The fine-tuned QMIX achieves significantly higher performance than its non-fine-tuned counterpart. Moreover, the fine-tuned QPLEX with random noise and zero vector outperforms both its stateful version and its non-tuned runs. The average return during training for these additional experiments is in \Cref{suppl:plots}. 

%% file: Sections/conclusions.tex
\section{Conclusions}
\label{sec:conclusions}

This work effectively addressed an important gap in the theoretical and practical underpinnings of value function factorization. We analyzed the relationship between theoretical frameworks and practical implementations and proposed a novel efficient factorization algorithm. 

From a theoretical standpoint, we formally analyze the mismatch between the stateless theoretical framework presented in prior research and the actual stateful algorithms. Our experiments further questioned the conventional use of the state during the factorization process. Contrary to common practice, where the state is employed as centralized information, our results suggest there could be better solutions. In particular, even simplistic forms of centralized random noise or zero vectors, exhibit comparable or superior performance compared to stateful algorithms in certain scenarios.

On the practical front, 
we introduced DuelMIX, a factorization scheme designed to learn more general and distinct per-agent utility estimators. Furthermore, DuelMIX combines history values in a weighted manner to refine the estimation of the joint history value. Experiments on StarCraft II micromanagement and complex coordination tasks demonstrate the benefits of our intuitions.

Our empirical insights not only contribute significantly to the current understanding of MARL but also lay a principled foundation for future research in this domain.

%% file: Sections/appendices.tex
\appendix
\onecolumn

\section{Value Factorization Method Variants}
\label{sec:methods}

In this section, we formalize variants of QMIX and QPLEX that use different types of information in their key components, e.g., nothing, joint history, state, or joint history and state.
Each variant employs per-agent utilities $Q_i(h_i, u_i)$ and combines them (optionally integrating joint history and/or state information) to obtain a joint action value function that takes the form of $Q(\bm{h}, \bm{u})$ for stateless variants, and $Q(\bm{h}, s, \bm{u})$ for stateful variants.

\subsection{QMIX}
\label{sec:qmix}

\begin{figure}[b]
    \centering
    \includegraphics[width=\linewidth]{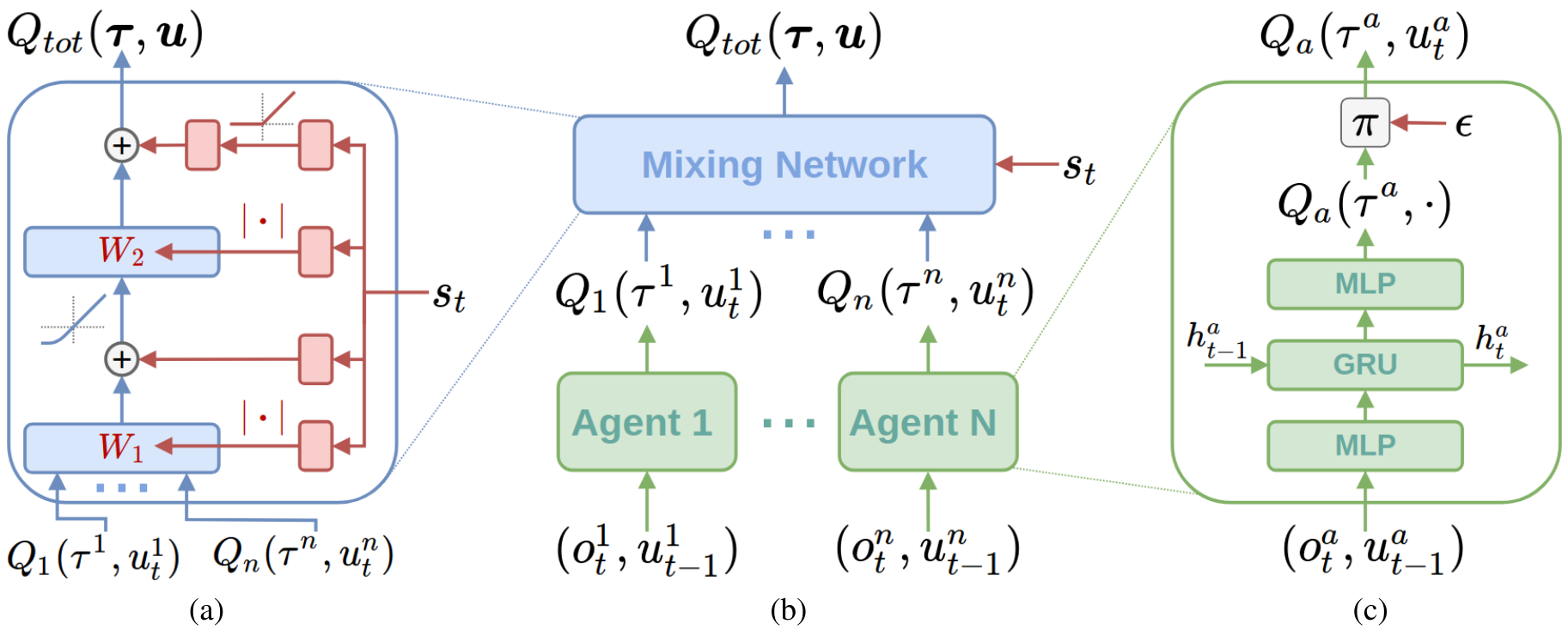}
    \caption{State-QMIX architecture (image credit: \cite{qmix}). (a) Mixing network; (b) QMIX architecture; (c) Individual utility networks. Note the use of state in the mixing network, which means that the output should more correctly be denoted as $Q_{tot}(\bm{\tau}, s, \bm{u})$. This discrepancy between theory and implementation may undermine the validity of the IGM principle  to this practical implementation of QMIX.}
    \label{fig:qmix}
\end{figure}

We categorize the QMIX variants in terms of which auxiliary input is provided to the mixing hyper-network in addition to the individual agent utilities.
In that regard, QMIX~\citep{qmix} is originally formalized directly in a state variant, while the IGM principle is formalized without any sort of auxiliary information. This discrepancy is not directly addressed, and may potentially undermine the validity of the theoretical guarantees for the practical implementation.

\subsubsection{Plain-QMIX}
\label{sec:history-qmix}

The Plain-QMIX variant is a direct implementation of the IGM principle, with no auxiliary information provided to set the hyper-network weights.
This variant uses a mixing network to combine the per-agent utilities $\langle Q_i(h_i, u_i) \rangle_{i\in\mathcal{N}}$ into a joint history-action value function $Q(\bm{h}, \bm{u}) = f(Q_1(h_1, u_i), \ldots, Q_n(h_n, u_n))$ that satisfies monotonic constraints between the joint $Q$ and the individual $Q_i$.
This variant trivially satisfies the IGM principle.

\subsubsection{State-QMIX}
\label{sec:state-qmix}

The State-QMIX variant uses ground truth state information to set the hyper-network weights, and it is the main variant proposed in its respective paper (see \Cref{fig:qmix}).
This variant uses a mixing \emph{hyper}-network to combine the per-agent utilities $\langle Q_i(h_i, u_i) \rangle_{i\in\mathcal{N}}$ and the ground truth state into a joint history-action value function $Q(\bm{h}, s, \bm{u}) = f(Q_1(h_1, u_i), \ldots, Q_n(h_n, u_n), s)$ that satisfies monotonic constraints between the joint $Q$ and the individual $Q_i$.
The original work by \cite{qmix} performs an empirical evaluation concerning the role of state. However, it does not formally demonstrate that a stateful mixing model still satisfies the stateless IGM principle.

\subsection{QPLEX}
\label{sec:qplex}

\begin{figure}
    \centering
    \includegraphics[width=\linewidth]{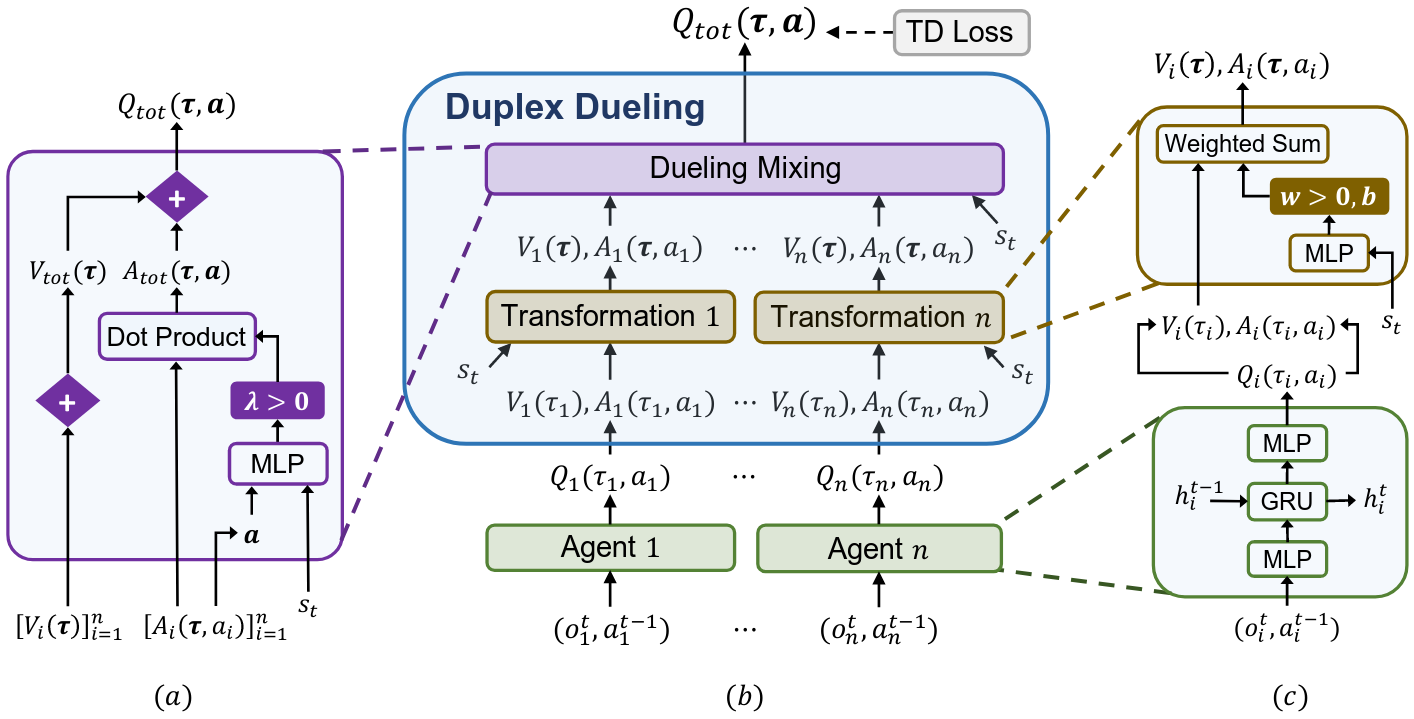}
    \caption{State-QPLEX architecture (image credit: \cite{qplex}). (a) Mixing network; (b) QPLEX architecture; (c) Individual utility and transformation networks. Note the use of state in the mixing and transformation networks, which means that the output should more correctly be denoted as $Q_{tot}(\bm{\tau}, s, \bm{u})$. This discrepancy between theory and implementation undermines the validity of \citet[Proposition~2]{qplex}, i.e., the action-value function class that the practical implementation of QPLEX can realize is not equivalent to what is induced by the IGM principle.}
    \label{fig:qplex}
\end{figure}

We categorize the QPLEX variants in terms of the inputs to the weight, bias, and attention models.
In that regard, the derivation and implementation of QPLEX~\citep{qplex} differ in ways that undermine the validity of the theoretical guarantees for the practical implementation.
In the following subsections, we formalize each variant clearly, highlighting the differences. The discussion of why these differences matter can be found in the main document of this submission.
The history variant corresponds to the theoretical formalization of QPLEX, the state variant corresponds to its practical implementation, while the history-state variant is our attempt at bridging the gap between theory and practice.

\subsubsection{History-QPLEX}
\label{sec:history-qplex}

History-QPLEX corresponds to the theoretical stateless formulation derived in the corresponding paper.
It decomposes the per-agent utilities $\langle Q_i(h_i, u_i) \rangle_{i\in\mathcal{N}}$ into individual values $\langle V_i(h_i) \rangle_{i\in\mathcal{N}}$ and advantages $\langle A_i(h_i, u_i) \rangle_{i\in\mathcal{N}}$ according to:
\begin{align}
    V_i(h_i) &= \max_{u_i'} Q_i(h_i, u_i'), &
    A_i(h_i, u_i) &= Q_i(h_i, u_i) - \max_{u_i'} Q_i(h_i, u_i').
\label{eq:history-qplex:values}
\end{align}

Such values are then conditioned on the joint history using a monotonic transformation. This module takes as input the joint history $\bm{h}$ and outputs biases and positive weights $\langle b_i(\bm{h}), w_i(\bm{h}) > 0 \rangle_{i\in\mathcal{N}}$:
\begin{align}
    V_i(\bm{h}) &= w_i(\bm{h}) V_i(h_i) + b_i(\bm{h}), &
    A_i(\bm{h}, u_i) &= w_i(\bm{h}) A_i(h_i, u_i). 
\label{eq:history-qplex:transformation}
\end{align}

After this transformation, History-QPLEX factorizes the joint value function as:
\begin{equation}
    Q(\bm{h}, \bm{u}) = \sum_i V_i(\bm{h}) + \lambda_i(\bm{h}, \bm{u}) A_i(\bm{h}, u_i),
\label{eq:history-qplex:joint}
\end{equation}
where $\langle \lambda_i(\bm{h}, \bm{u}) > 0 \rangle_{i\in\mathcal{N}}$ are weights computed with an attention module to enhance credit assignment.
Both the transformation and the attention module use positive weights to maintain action selection consistency over the advantage values. Naturally, this positivity enforces monotonicity in the advantage factorization.

\subsubsection{State-QPLEX}
\label{sec:state-qplex}

State-QPLEX corresponds to the stateful practical implementation used in the empirical evaluation in the corresponding paper (see \Cref{fig:qplex}). The main difference compared to History-QPLEX is that the transformation and attention modules replace the joint history for the underlying ground truth state.

State-QPLEX decomposes the per-agent utilities $\langle Q_i(h_i, u_i) \rangle_{i\in\mathcal{N}}$ into individual values $\langle V_i(h_i) \rangle_{i\in\mathcal{N}}$ and advantages $\langle A_i(h_i, u_i) \rangle_{i\in\mathcal{N}}$ according to:

\begin{align}
    V_i(h_i) &= \max_{u_i'} Q_i(h_i, u_i'), &
    A_i(h_i, u_i) &= Q_i(h_i, u_i) - \max_{u_i'} Q_i(h_i, u_i').
\label{eq:state-qplex:values}
\end{align}

Such values are then conditioned on the ground truth state using a monotonic transformation. This module takes as input the state $s$ and outputs biases and positive weights $\langle b_i(s), w_i(s) > 0 \rangle_{i\in\mathcal{N}}$:
\begin{align}
    V_i(h_i, s) &= w_i(s) V_i(h_i) + b_i(s), &
    A_i(h_i, s, u_i) &= w_i(s) A_i(h_i, u_i). 
\label{eq:state-qplex:transformation}
\end{align}

After this transformation, State-QPLEX factorizes the joint value function as:
\begin{equation}
    Q(\bm{h}, s, \bm{u}) = \sum_i V_i(h_i, s) + \lambda_i(s, \bm{u}) A_i(h_i, s, u_i),
\label{eq:state-qplex:joint}
\end{equation}
where $\langle \lambda_i(s, \bm{u}) > 0 \rangle_{i\in\mathcal{N}}$ are weights computed with an attention module to enhance credit assignment.
Note that the joint value function is now also a function of state, due to the dependence on the stateful transformation and attention modules.

\subsubsection{History-State-QPLEX}
\label{sec:history-state-qplex}

History-State-QPLEX is our proposed attempt at unifying the stateless theoretical derivation with the stateful practical implementation. The main difference compared to previous variants is that the transformation and attention modules employ \emph{both} the joint history and the underlying ground truth state.

State-QPLEX decomposes the per-agent utilities $\langle Q_i(h_i, u_i) \rangle_{i\in\mathcal{N}}$ into individual values $\langle V_i(h_i) \rangle_{i\in\mathcal{N}}$ and advantages $\langle A_i(h_i, u_i) \rangle_{i\in\mathcal{N}}$ according to:
\begin{align}
    V_i(h_i) &= \max_{u_i'} Q_i(h_i, u_i'), &
    A_i(h_i, u_i) &= Q_i(h_i, u_i) - \max_{u_i'} Q_i(h_i, u_i').
\label{eq:history-state-qplex:values}
\end{align}

Such values are then conditioned on the joint history and the ground truth state using a monotonic transformation. This module takes as input the joint history $\bm{h}$ and state $s$ and outputs biases and positive weights $\langle b_i(\bm{h}, s), w_i(\bm{h}, s) > 0 \rangle_{i\in\mathcal{N}}$:
\begin{align}
    V_i(\bm{h}, s) &= w_i(\bm{h}, s) V_i(h_i) + b_i(\bm{h}, s), &
    A_i(\bm{h}, s, u_i) &= w_i(\bm{h}, s) A_i(h_i, u_i). 
\label{eq:history-state-qplex:transformation}
\end{align}

After this transformation, History-State-QPLEX factorizes the joint value function as:
\begin{equation}
    Q(\bm{h}, s, \bm{u}) = \sum_i V_i(\bm{h}, s) + \lambda_i(\bm{h}, s, \bm{u}) A_i(\bm{h}, s, u_i),
\label{eq:history-state-qplex:joint}
\end{equation}
where $\langle \lambda_i(\bm{h}, s, \bm{u}) > 0 \rangle_{i\in\mathcal{N}}$ are weights computed with an attention module to enhance credit assignment.
As in the previous case, the joint value function is now also a function of state, due to the dependence on the stateful transformation and attention modules.

\section{Stateful Value Factorization Proofs}
\label{sec:methods_analysis:proof}

In this section, we formally present the proofs for \Cref{thm:qmix-state-bias,thm:qplex-state-bias,thm:duelmix-state-bias}.

A common theme among all methods and proofs is the conclusion that the constraints that guarantee the mixing model's monotonicity are so restrictive that they intrinsically do not allow state information to alter the identity of the maximal action, leading to the same actions being taken during both training and execution.  While the state does influence the values that are being learned, actions are always obtained in a way that is intrinsically consistent with the stateless context of partial observability.
This is not to say that the methods do not present any other type of bias; in fact, it would be more accurate to say that the modeling bias of the mixing model, which specifically prevents it from becoming a universal function approximator, expands widely enough to subsume any possible bias associated with the use of the state.

\subsection{Proof of \Cref{thm:qmix-state-bias}}

\begin{proposition*}[QMIX State Bias]
A stateful implementation of QMIX does not introduce any additional state-induced bias compared to a stateless implementation.
\end{proposition*}

We begin by adjusting the notation of centralized value models to show the use of state, effectively resulting in history-state values $Q(\bm{h}, s, \bm{u})$.  Given that consistent history and history-state values are related by $Q(\bm{h}, \bm{u}) = \mathbb{E}_{s\mid \bm{h}}\left[ Q(\bm{h}, s, \bm{u}) \right]$, the IGM principle is equivalently written in a marginalized history-state form as follows:
\begin{equation}
    \argmax_{\bm{u} \in \mathcal{U}} \mathbb{E}_{s\mid \bm{h}}\left[ Q(\bm{h}, s, \bm{u}) \right] \equiv 
    \left(
    \argmax\limits_{u_i \in U_i}Q_i(h_i, u_i)
    \right)_{i\in\mathcal{N}}.
\label{eq:igm:Ehs:appendix}
\end{equation}

The question now becomes whether the mixing model of QMIX satisfies this extended History-State-IGM principle.  Given QMIX's hyper-network architecture and the way that state is used, the monotonicity constraint holds even for history-state values, resulting in a \emph{non-marginalized} version of History-State-IGM:
\begin{equation}
    \argmax_{\bm{u} \in \mathcal{U}} Q(\bm{h}, s, \bm{u}) \equiv 
    \left(
    \argmax\limits_{u_i \in U_i}Q_i(h_i, u_i)
    \right)_{i\in\mathcal{N}}.
\label{eq:igm:hs:appendix}
\end{equation}

While \Cref{eq:igm:hs:appendix} is distinct from \Cref{eq:igm:Ehs:appendix}, we note that the QMIX architecture forces this non-marginalized version of IGM to hold for any state, i.e., different states will result in different centralized values, but the identity of the maximal action is invariant to the state itself.  In turn, this implies that the maximal action associated with \emph{any} marginalization over states of the centralized model remains invariant, as well, i.e., for any $p\in\Delta\mathcal{S}$:
\begin{equation}
    \argmax_{\bm{u} \in \mathcal{U}} \mathbb{E}_{s\sim p} \left[ Q(\bm{h}, s, \bm{u}) \right] \equiv 
    \left(
    \argmax\limits_{u_i \in U_i}Q_i(h_i, u_i)
    \right)_{i\in\mathcal{N}}.
\end{equation}

As this relationship holds for any distribution over states, it naturally holds specifically for the joint belief $\Pr(s\mid \bm{h})$, which directly implies the marginalized History-State IGM principle
of \Cref{eq:igm:Ehs:appendix}.

\subsection{Proof of \Cref{thm:qplex-state-bias}}

The analysis for QPLEX follows broadly a similar structure to that for QMIX.  We begin by adjusting the model notation to include state, reformulating \Cref{eq:qplex_transf} as:
\begin{equation}
\begin{split}
    &V_i(\bm{h}, s) = w_i(\bm{h}, s)V_i(h_i)+b_i(\bm{h}, s), \\
    &A_i(\bm{h}, s, u_i) = w_i(\bm{h}, s)A_i(h_i, u_i), 
\label{eq:qplex_transf:hs:appendix}
\end{split}
\end{equation}
\noindent and \Cref{eq:qplex_joint} as:
\begin{align}
    Q(\bm{h}, s, \bm{u}) &= V(\bm{h}, s) + A(\bm{h}, s, \bm{u}), \label{eq:qplex_joint:hs:appendix:q} \\
    \intertext{where}
    V(\bm{h}, s) &= \sum_i V_i(\bm{h}, s), \label{eq:qplex_joint:hs:appendix:v} \\
    A(\bm{h}, s, \bm{u}) &= \sum_i \lambda_i(\bm{h}, s, \bm{u})A_i(\bm{h}, s, u_i). \label{eq:qplex_joint:hs:appendix:a}
\end{align}

Notably, the structure of this stateful QPLEX establishes a connection that starts from the individual $Q_i(h_i, u_i)$, passes through $A_i(h_i, u_i)$, $A_i(\bm{h}, s, u_i)$, and $A(\bm{h}, s, \bm{u})$, and finally terminates at the centralized $Q(\bm{h}, s, \bm{u})$.  Further, each of these connections is constrained to not allow the state information to alter the identity of the maximal action.
We note that the architectural constraints that guarantee the Advantage-IGM principle in the stateless case are now guaranteeing an analogous \emph{non-marginalized} History-State Advantage-IGM principle:
\begin{align}
    \argmax_{\bm{u} \in \mathcal{U}} ~A(\bm{h}, s, \bm{u})
    &\equiv  \left( \argmax\limits_{u_i \in U_i}A(\bm{h}, s, u_i) \right)_{i\in\mathcal{N}} \nonumber \\
    &\equiv  \left( \argmax\limits_{u_i \in U_i}A_i(h_i, s, u_i) \right)_{i\in\mathcal{N}} \nonumber \\
    &\equiv  \left( \argmax\limits_{u_i \in U_i}A_i(h_i, u_i) \right)_{i\in\mathcal{N}}, \label{eq:app:hs:aigm}
\end{align}
\noindent and therefore the \emph{non-marginalized} History-State IGM principle:
\begin{align}
    \argmax_{\bm{u} \in \mathcal{U}} ~Q(\bm{h}, s, \bm{u})
    &\equiv \argmax_{\bm{u} \in \mathcal{U}} ~A(\bm{h}, s, \bm{u}) \nonumber \\
    &\equiv  \left( \argmax\limits_{u_i \in U_i}A_i(h_i, u_i) \right)_{i\in\mathcal{N}} \nonumber \\
    &\equiv  \left( \argmax\limits_{u_i \in U_i}Q_i(h_i, u_i) \right)_{i\in\mathcal{N}}. \label{eq:app:hs:igm}
\end{align}

As in the proof for QMIX, we conclude that if the maximal action associated with a centralized history-state value is invariant to changes of the state, then it is also invariant to any marginalization of the state, resulting in the marginalized History-State IGM principle.

\subsection{Proof of \Cref{thm:duelmix-state-bias}}

The structure of DuelMIX most closely resembles that of QPLEX. In relation to the use of state, two relevant changes must be inspected to prove the lack of state-induced bias.  In both cases, it is easy to show that the analysis for QPLEX applies all the same to DuelMIX.

\paragraph{Dueling Value Decomposition}
Instead of learning a single individual $Q_i(h_i, u_i)$ per agent that is then decomposed into $V_i(h_i)$ and $A(h_i, u_i)$, DuelMIX instead learns the two values separately (albeit with some parameter/model sharing).  Notably, this dueling value decomposition of the individual models does not depend on the state to begin with, and therefore cannot influence the method's bias concerning the use of state.

\paragraph{Weighting Value Mixing}
Compared to QPLEX, the weighted value mixing of \Cref{eq:duelmix_joint_v} introduces another point-of-entry for state information. However, it only influences the centralized value $V(\bm{h}, s)$ and none of the advantage values.  Therefore, it has no impact on the identity of the maximal action, and the relationships from \Cref{eq:app:hs:aigm,eq:app:hs:igm} continue to hold.

\newpage

\section{Proof of DuelMIX Expressiveness}
\label{sec:duelmix_proof}
\begin{proposition*}
    The function class that DuelMIX can realize is equivalent to what is induced by History-State Advantage-IGM.
\end{proposition*}

The proof follows the one provided by \citet{qplex}. The main differences are: (i) the transformation module estimates an additional bias term $c_i(s, \bm{u})$ for the local advantage utilities, and (ii) the joint observation history value is a non-linear combination of the local transformations and weights $w_i'(\bm{h}, s)$, which is the key component allowing DuelMIX to achieve full expressiveness.

\begin{proof}
    Assume DuelMIX's network size is enough to satisfy the universal function approximation theorem \citep{univ_approx}. 
    
    Denote the joint $Q^\text{MIX}, V^\text{MIX}, A^\text{MIX}$, transformed $Q^\text{T-MIX}_i, V^\text{T-MIX}_i, A^\text{T-MIX}_i$ and individual $Q_i^\text{MIX}, V_i^\text{MIX}, A_i^\text{MIX}$ action, observation, advantage history-based value functions learned by DuelMIX, respectively. Moreover, let the class of action-value functions that DuelMIX can represent be $\bm{\mathcal{Q}}^\text{MIX}$ defined as:
    \begin{equation}
    \begin{split}
    \bm{\mathcal{Q}}^\text{MIX} = \biggl\{ 
        &\left( Q^\text{MIX}:\mathcal{H}\times\mathcal{S}\times\mathcal{U}\rightarrow\mathbb{R}, \langle Q_i^\text{MIX}:H_i\times\mathcal{S}\times U_i\rightarrow\mathbb{R}\rangle_{i \in \mathcal{N}}
        \right) \\
        & \mid\text{\Cref{eq:qplex_transf:hs,eq:duelmix_joint_a,eq:duelmix_joint_v,eq:duelmix_jointq} are satisfied}\biggr\},
    \end{split}
    \end{equation}
    and let $\bm{\mathcal{Q}^{\text{IGM}}}$ be the class of action-value functions represented by IGM:
    \begin{equation}
    \begin{split}
        \bm{\mathcal{Q}}^\text{IGM} = \biggl\{ 
            &\left( Q^{\text{IGM}}:\mathcal{H}\times\mathcal{S}\times\mathcal{U}\rightarrow\mathbb{R}, \langle Q_i^{\text{IGM}}:H_i\times\mathcal{S}\times U_i\rightarrow\mathbb{R}\rangle_{i \in \mathcal{N}}
            \right) \\
            &\mid \text{\Cref{eq:igm:hs} holds} \biggr\}.
    \end{split}
    \end{equation}
    First, we note the multiplicative weights to the advantage function are all positive to satisfy action selection consistency. 
    
    We have to prove $\bm{\mathcal{Q}}^\text{MIX} \equiv \bm{\mathcal{Q}}^{\text{IGM}}$. To this end, we demonstrate the inclusion in the two directions.

    \begin{enumerate}
        \item $\bm{\mathcal{Q}}^{\text{IGM}} \subseteq \bm{\mathcal{Q}}^\text{MIX}$: for any $\left(Q^{\text{IGM}}, \langle Q_i^{\text{IGM}}\rangle_{i\in\mathcal{N}}\right)\in\mathcal{Q}^{\text{IGM}}$ we construct $Q^\text{MIX}=Q^{\text{IGM}}$ and $\langle Q_i^\text{MIX}\rangle_{i\in\mathcal{N}} = \langle Q_i^{\text{IGM}}\rangle_{i\in\mathcal{N}}$. 

        By recalling the construction of DuelMIX in \Cref{eq:qplex_transf:hs,eq:duelmix_jointq}, and setting:
        \begin{equation*}
        \begin{split}            
            &w_i(s) = w_i'(\bm{h}, s) = 1, \\ &b_i(s) = V_i^\text{T-MIX}(h_i, s)-V_i^\text{MIX}(h_i), ì \\
            &\lambda_i(s, \bm{u}) = \begin{cases}
                1 & \text{if}~A_i^\text{MIX}(h_i, u_i) = 0 \\
                \frac{A_i^\text{T-MIX}(h_i, s, u_i)}{A_i^\text{MIX}(h_i, u_i)} > 0 & \text{otherwise}
            \end{cases}
        \end{split}
        \end{equation*}
        we can conclude $\left(Q^\text{MIX}, \langle Q_i^\text{MIX}\rangle_{i\in\mathcal{N}}\right)\in\mathcal{Q}^\text{MIX}$, meaning that $\bm{\mathcal{Q}}^{\text{IGM}}\subseteq\bm{\mathcal{Q}}^\text{MIX}$.

        \item $\bm{\mathcal{Q}}^{\text{MIX}} \subseteq \bm{\mathcal{Q}}^\text{IGM}$: for any $\left(Q^{\text{MIX}}, \langle Q_i^{\text{MIX}}\rangle_{i\in\mathcal{N}}\right)\in\mathcal{Q}^{\text{MIX}}$, let $u_i^*$ be the maximizing action for the individuals advantage:
        \begin{equation*}
            u_i^* = \argmax_{u\in U_i}A_i^{\text{MIX}}(h_i, u_i)
        \end{equation*}
        Following DuelMIX's architecture and, in particular, the per-agent advantage output stream $A_i^\text{MIX}(h_i,\cdot) - \max_{u_i} A_i^\text{MIX}(h_i,u_i)$, we can derive $\forall i \in \mathcal{N}, u_i \neq u_i^* \in U_i$:
        \begin{equation*}
            A_i^{\text{MIX}}(h_i, u_i^*) = 0 ~ \wedge ~ A_i^{\text{MIX}}(h_i, u_i) < 0 \qquad \text{\small{(Per-agent utilities)}}
        \end{equation*}
        \vspace{5pt}
        \begin{equation*}
        \begin{split}
            \implies &A_i^{\text{T-MIX}}(h_i, s, u_i^*) = w_i(s)A_i^{\text{MIX}}(h_i, u_i^*) = 0 \qquad \text{\small{(Transformation)}}\\
            &\wedge ~ A_i^{\text{T-MIX}}(h_i, s, u_i) = w_i(s)A_i^{\text{MIX}}(h_i, u_i) < 0 
        \end{split}
        \end{equation*}
        \vspace{5pt}
        \begin{equation*}
        \begin{split}
            \implies &A_i^{\text{MIX}}(\bm{h}, s, \bm{u}^*) = \sum_{i}^{\mathcal{N}}\lambda_i(s,\bm{u}^*)A_i^{\text{T-MIX}}(h_i, s, u_i^*) = 0 \qquad \text{\small{(Mixing)}} \\
            &\wedge ~ A_i^{\text{MIX}}(\bm{h}, s, \bm{u}) = \sum_{i}^{\mathcal{N}}\lambda_i(s,\bm{u})A_i^{\text{T-MIX}}(h_i, s, u_i) < 0
        \end{split}
        \end{equation*}
        We can thus form the set of maximizing joint actions in terms of the maximizing individuals' actions and construct $Q^\text{IGM}=Q^{\text{MIX}}$ and $\langle Q_i^\text{IGM}\rangle_{i\in\mathcal{N}} = \langle Q_i^{\text{MIX}}\rangle_{i\in\mathcal{N}}$. According to these, the action-selection consistency of History-State Advantage-IGM is satisfied. We can conclude $\left(Q^\text{IGM}, \langle Q_i^\text{IGM}\rangle_{i\in\mathcal{N}}\right)\in\mathcal{Q}^\text{IGM}$, meaning that $\bm{\mathcal{Q}}^{\text{MIX}}\subseteq\bm{\mathcal{Q}}^\text{IGM}$.
    \end{enumerate}
\end{proof}

\newpage
\section{Hyper-parameters}
\label{suppl:hyperparameters}
Regarding the considered baselines, we employed the original authors' mixer implementations and most parameters \citep{vdn, qmix, qplex}.
Table \ref{tab:gridsearch} lists the key hyper-parameters considered in our initial grid search for tuning the algorithm employed in Section \ref{sec:experiments}, highlighting the best one in the last column. When indicating two best values, the first one is employed for Box Pushing, and the other in the SMACLite domains.
\begin{table}[h]
\centering
\caption{Hyper-parameters candidate for initial grid search tuning, and best parameters.}
\label{tab:gridsearch}
\begin{tabular}{llll}
\toprule
\textbf{Module}         & \textbf{Parameter}           & \textbf{Grid Search}           & \textbf{Best Values}  \\ \midrule
\textbf{Algorithm}      & Learning rate       & 1e-3, 2.5e-4, 1e-4    & 2.5e-4, 1e-4 \\
               & $\gamma$            & 0.9, 0.95, 0.97, 0.99 & 0.9, 9.77    \\
               & Max-grad norm       & 10, 40, 80            & 10           \\
               & Buffer size         & 10000, 50000, 500000  & 10000, 50000 \\
               & Batch size          & 32, 64, 128           & 64           \\
               & Target update freq. & 1000, 2500, 5000      & 2500, 5000   \\ \midrule
\textbf{Agent network}  & N° layers           & 1, 2                  & 2            \\
               & Size                & 64, 128, 256          & 128          \\ \midrule
\textbf{Mixing modules} & N° layers           & 1, 2                  & 2            \\
               & Size                & 32, 64                & 32           \\
               & N° attention heads  & 4, 8                  & 4            \\ \bottomrule
\end{tabular}
\end{table}

\section{Environmental Impact}
\label{suppl:env_impact}
Despite being significantly less computationally demanding than the original SMAC, the high number of experiments in SMACLite led to environmental impacts due to intensive computations that (possibly) run on computer clusters for an extended time. Our experiments were conducted using a private infrastructure with a carbon efficiency of $\approx 0.275 \frac{kgCO_2eq}{kWh}$, requiring a cumulative $\approx$170 hours of computation. Total emissions are estimated to be $\approx3.97 kgCO_2eq$ using the \href{https://mlco2.github.io/impact#compute}{Machine Learning Impact calculator} of which 100$\%$ were offset through \href{https://www.treedom.net}{Treedom}.

\section{Environments}
\label{sec:environments}
\begin{figure}[h]
    \centering
    \includegraphics[width=.5\linewidth]{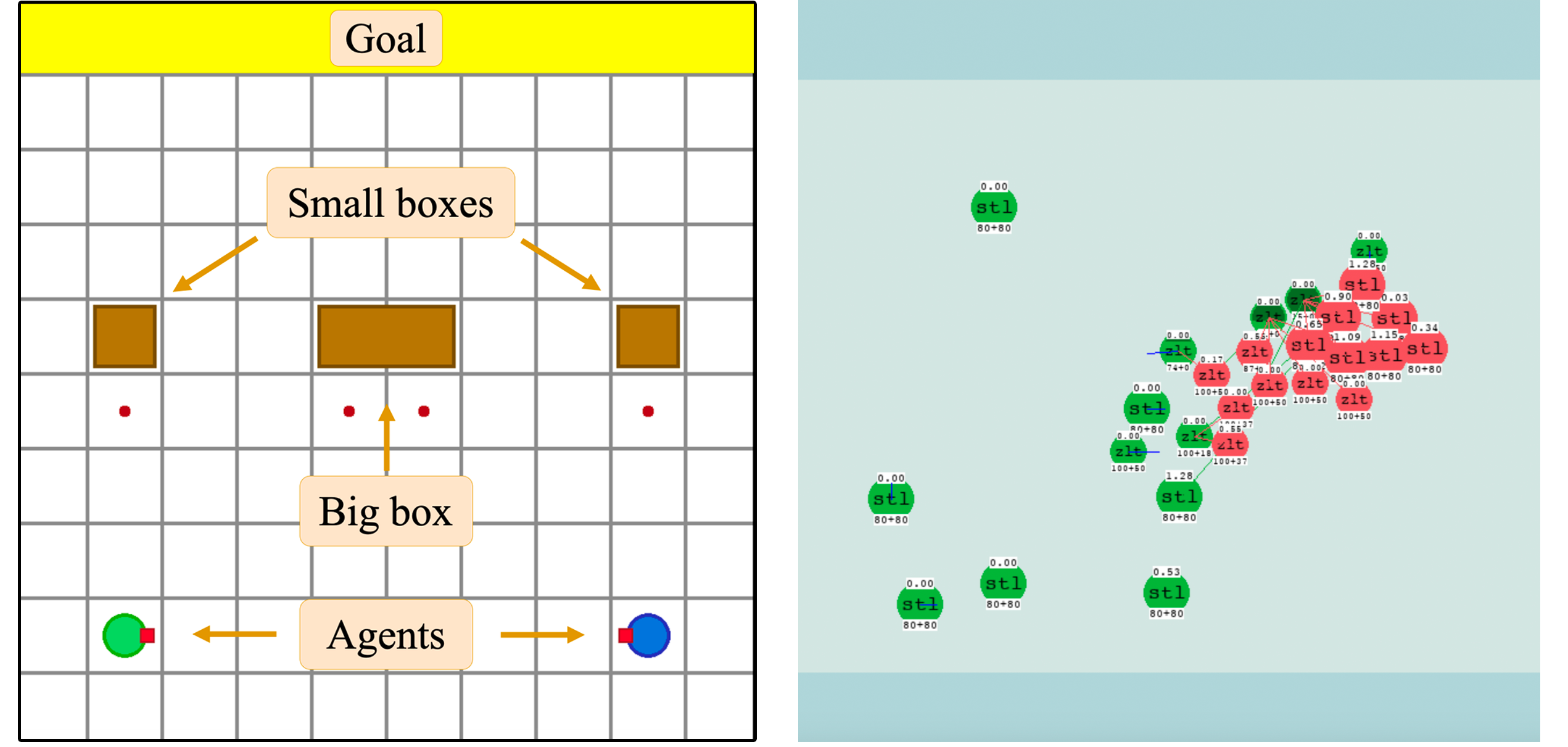}
    \caption{Representative overview of the SMACLite 7sz task (left) and Box Pushing (right).}
    \label{fig:scenarios}
\end{figure}

\Cref{fig:scenarios} shows a representative game view of Box Pushing on the left, and SMACLite on the right.

\subsection{Box Pushing 30}
In this collaborative task, two agents have to work together to push a big box to a goal area at the top of a grid world of size 30 to obtain a higher credit than pushing the small box on each own. The small box is movable with a single agent, while the big one requires two agents to push it simultaneously. 

The state space consists of each agent's position and orientation, as well as the location of each box. Agents have a set of discrete actions, including \textit{moving forward}, \textit{turning left} or \textit{right}, and \textit{staying} in place. Each agent observation is very limited as it only observes the state of the front cell: empty, teammate, boundary, small box, or big box.

The team receives a terminal reward of $+100$ for pushing the big box to the goal area or $+10$ for pushing one small box to the goal area. If any agent hits the world's boundary or pushes the big box on its own, a penalty of $-10$ is issued. An episode terminates when any box is moved to the goal area or reaches the maximum horizon, 100 time steps.

BP represents a significantly harder task compared to SMACLite maps. Optimal behaviors are particularly challenging to discover and learn, requiring tight coordination under severe partial observability. Specifically, agents should both face the big box at the same time, check whether the other agent is next to them, and then constantly push the box.

\subsection{StarCraft II Lite}
SMACLite \citep{smaclite} has been recently introduced as a computationally lightweight alternative to StarCraft II unit micromanagement tasks \citep{starcraft}. The lite version of the benchmark considers the same decentralized combat scenarios, where enemy units are controlled by an AI, and each allied unit is a reinforcement learning agent. Crucially, \citet{smaclite} showed that policies trained in SMACLite are directly transferable to the original SMAC benchmarks. This confirms that both suites of tasks represent comparable challenges. The main difference lies in the evaluation metrics. While SMAC typically employs the average win rate of the MARL team, SMACLite uses the average return which is typical in other reinforcement learning and MARL benchmarks. In particular, a team of agents can achieve a normalized return in the range $[0, 20]$, where 0 indicates no damage was inflicted on the enemies, and 20 means the MARL team successfully defeats the enemies. As such, 20 return maps to 100\% win rate for the RL agents.

At every step, agents sample actions from a discrete space to perform a movement direction (in the continuous map), attacking a certain enemy in their shooting range, stopping, or doing nothing. Agents get a joint reward based on the damage inflicted on the enemies, with a sparse value of 10 and 200 for killing a unit and winning the combat, respectively. Our experiments, summarized in Table 4, consider both standard maps, as well as recent super-hard tasks introduced by QPLEX \citep{qplex}. 

We refer to the original papers \citep{smaclite, starcraft} for a detailed discussion about the state of the environment and agents' observations.

\begin{table}[h]
\centering
\label{tab:smac}
\caption{SMACLite tasks considered in our experiments.}
\begin{tabular}{llll}
\toprule
\textbf{Map Name}  & \textbf{Difficulty} & \textbf{Ally Units}               & \textbf{Enemy Units}              \\ \midrule
3s5z               & \textit{Easy}       & 3 Stalkers, 3 Zealots             & 3 Stalkers, 3 Zealots             \\ \midrule
10m\_vs\_11m        & \textit{Hard} & 10 Marines                         & 11 Marines                   \\
3s8z\_vs\_3s9z & \textit{Hard} & 3 Stalkers, 8 Zealots & 3 Stalkers, 9 Zealots \\
27m\_vs\_30m        & \textit{Hard} & 27 Marines                         & 30 Marines                                         \\ \midrule
bane\_vs\_bane              & \textit{Super-hard} & 20 Zerglings, 4 Banelings            & 20 Zerglings, 4 Banelings            \\
5s10z              & \textit{Super-hard} & 5 Stalkers, 10 Zealots            & 5 Stalkers, 10 Zealots            \\
7sz                & \textit{Super-hard} & 7 Stalkers, 7 Zealots             & 7 Stalkers, 7 Zealots             \\ \bottomrule
\end{tabular}
\end{table}

\newpage
\section{Omitted Figures in \Cref{sec:experiments}}
\label{suppl:plots}

\begin{figure*}[h]
    \centering
      \includegraphics[width=.8\textwidth]{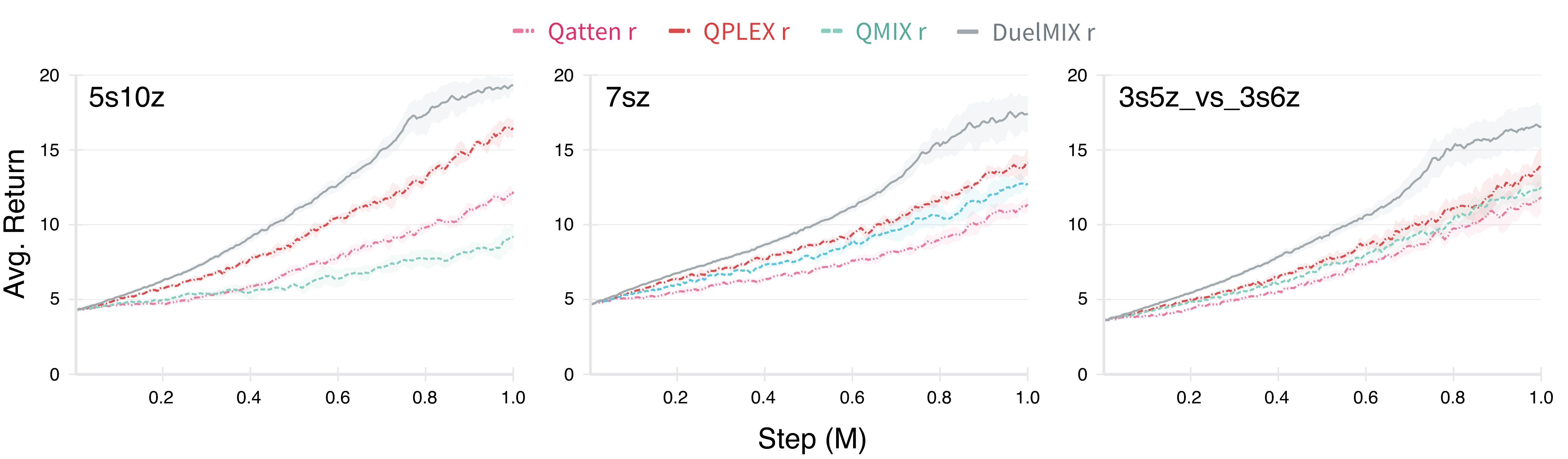}
      \includegraphics[width=.8\textwidth]{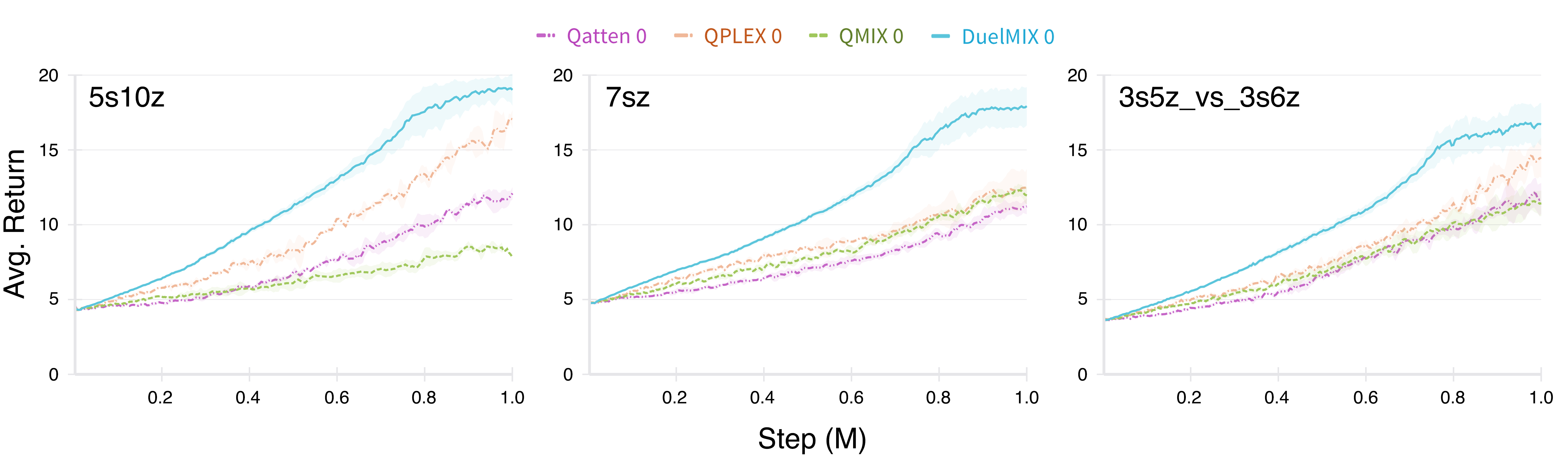}
      \caption{Average return during the training for factorization algorithms employing: (i) centralized random noise, marked with r (top row), and (ii) a zero vector, marked with 0 (bottom row).}
      \label{fig:res_stateinfo}
\end{figure*}

\Cref{fig:res_stateinfo} shows the average return during training of the stateful value factorization algorithms employing two types of centralized information: (i) uniform random noise sampled between $[0, 1.0]$, and (ii) a constant zero vector. 

\begin{figure}[h]
    \centering
    \includegraphics[width=0.6\linewidth]{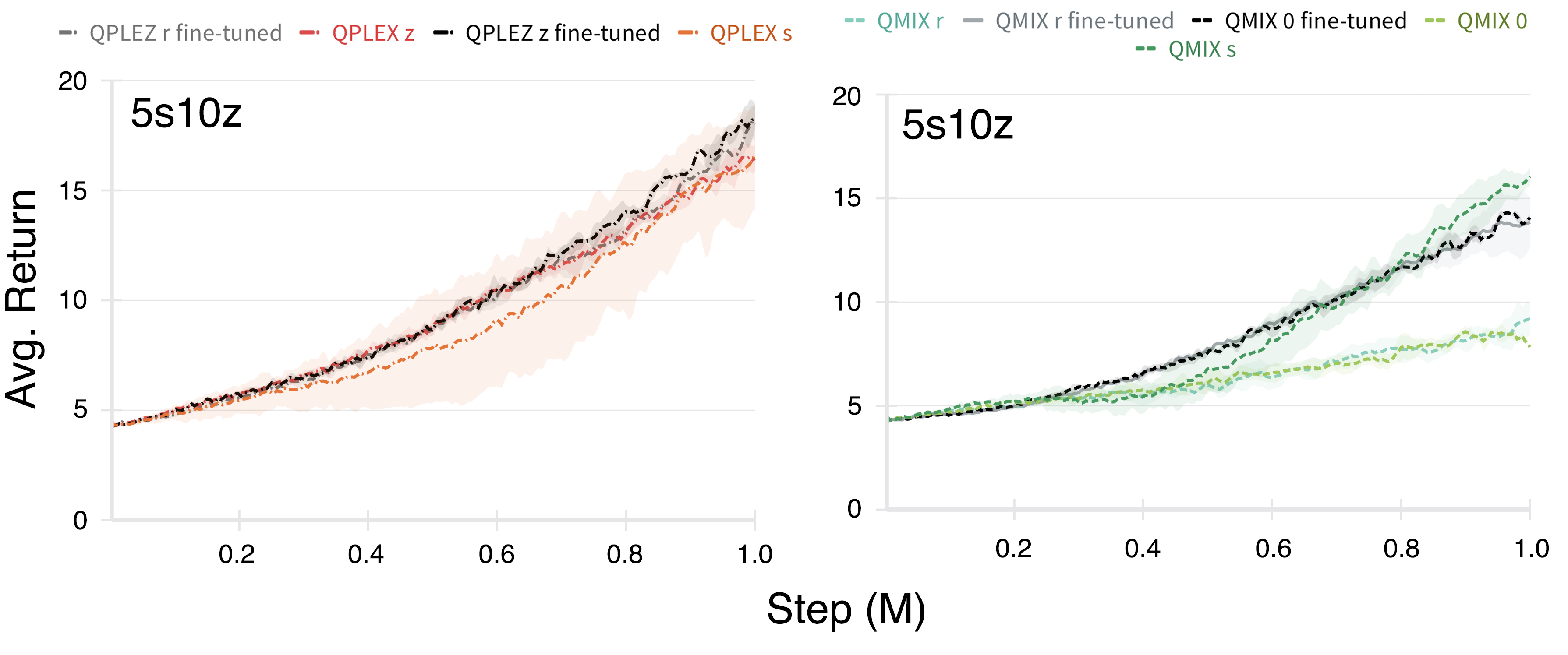}
    \vspace{-0.5cm}
    \caption{Learning curves for fine-tuned QPLEX (left) and QMIX (right) with random and zero centralized information.}
    \vspace{-0.3cm}
    \label{fig:res_finetune}
\end{figure}

\Cref{fig:res_finetune} extends the previous evaluation showing the performance of the fine-tuned QMIX and QPLEX using this alternative centralized information. Results for Qatten and DuelMIX are not included since additional fine-tuning did not significantly increase their performance.

\newpage
\section{Omitted State Results in \Cref{sec:experiments}}
\label{sec:state_results}

\begin{table}[h!]
    \setlength{\tabcolsep}{4pt}
    \centering
    \caption{Average convergence return when using centralized state (s), random noise (r), or constant vector (c).}
    \label{tab:res_stateinfo}
    \centering
    \begin{tabular}{lllll}
\toprule
        & \multicolumn{1}{c}{} & \textbf{3s5z\_vs\_3s6z}                                           & \textbf{5s10z}                                                    & \textbf{7sz}                                  \\ \midrule
\textbf{QMIX}    & \textit{s}           & \textit{16.0 $\pm$ 2.7} & \textit{15.8 $\pm$ 0.4} & \textit{16.7 $\pm$ 1.9} \\
        & \textit{r}           & 12.4 $\pm$ 0.9                              & 9.2 $\pm$ 0.8                      & 12.7 $\pm$ 0.3                     \\
        & \textit{0}           & 11.4 $\pm$ 0.9                              & 8.0 $\pm$ 0.2                               & 12.0 $\pm$ 0.6                              \\ \midrule
\textbf{Qatten}  & \textit{s}           & \textit{12.2 $\pm$ 0.4}                     & \textit{16.5 $\pm$ 1.7}                     & 10.7 $\pm$ 0.8                              \\
        & \textit{r}           & 11.8 $\pm$ 1.1                              & 12.0 $\pm$ 0.6                              & \textit{11.3 $\pm$ 0.5}                     \\
        & \textit{0}           & 11.7 $\pm$ 1.1                              & 11.9 $\pm$ 0.3                              & 11.2 $\pm$ 0.5                              \\ \midrule
\textbf{QPLEX}   & \textit{s}           & \textit{17.4 $\pm$ 2.7}                     & 16.4 $\pm$ 2.4                              & \textit{16.3 $\pm$ 3.2}                     \\
        & \textit{r}           & 13.9 $\pm$ 1.2                              & 16.4 $\pm$ 0.6                              & 14.1 $\pm$ 0.8                              \\
        & \textit{0}           & 14.4 $\pm$ 1.2                              & \textit{17.1 $\pm$ 0.3}                     & 12.5 $\pm$ 1.0                              \\ \midrule
\textbf{DuelMIX} & \textit{s}           & \textit{\textbf{18.6 $\pm$ 0.8}}            & 19.0 $\pm$ 0.7                              & \textit{\textbf{19.1 $\pm$ 0.4}}            \\
        & \textit{r}           & 17.2 $\pm$ 1.5                              & \textit{\textbf{19.3 $\pm$ 0.5}}            & 17.5 $\pm$ 1.3                              \\
        & \textit{0}           & 17.0 $\pm$ 1.1                              & 19.2 $\pm$ 0.7                              & 17.9 $\pm$ 1.3                              \\ \bottomrule
    \end{tabular}
\end{table}